\documentclass[10pt,twocolumn,letterpaper]{article}
\usepackage[pagenumbers]{cvpr} 
\usepackage{graphicx}
\usepackage{amsmath}
\usepackage{amssymb}
\usepackage{booktabs}
\usepackage{amsthm}
\usepackage{caption}
\usepackage{amsmath}
\usepackage[utf8]{inputenc} 
\usepackage[T1]{fontenc}    
\usepackage{url}            
\usepackage{booktabs}       
\usepackage{amsfonts}       
\usepackage{nicefrac}       
\usepackage{microtype}      
\usepackage{xcolor}         
\usepackage{subcaption}
\usepackage{caption}
\usepackage{amsmath}
\usepackage{amsthm}
\usepackage{multirow}
\usepackage{graphicx}
\usepackage[pagebackref,breaklinks,colorlinks]{hyperref}
\hypersetup{colorlinks,linkcolor={blue},citecolor={blue},urlcolor={blue}}  
\usepackage[capitalize]{cleveref}
\crefname{section}{Sec.}{Secs.}
\Crefname{section}{Section}{Sections}
\Crefname{table}{Table}{Tables}
\crefname{table}{Tab.}{Tabs.}
\usepackage{comment}
\usepackage{booktabs}
\usepackage{calc}
\usepackage{enumitem}
\renewcommand{\vec}[1]{\ensuremath{\mathbf{#1}}}
\newcommand{\dataset}{\ensuremath{\mathcal{D}}}
\newcommand{\datatrain}{\ensuremath{\mathcal{S}}}
\newcommand{\datatest}{\ensuremath{\mathcal{T}}}
\newcommand{\databatch}{\ensuremath{\mathcal{B}}}
\newcommand{\trainmean}{\ensuremath{\vec{\hat{\mu}}_l}}
\newcommand{\trainvar}{\ensuremath{\vec{\hat{\sigma}}_l^2}}

\newcommand\blfootnote[1]{%
  \begingroup
  \renewcommand\thefootnote{}\footnote{#1}%
  \addtocounter{footnote}{-1}%
  \endgroup
}

\newcommand{\mycomment}[1]{}

\newif\ifdraft
\draftfalse

\definecolor{orange}{rgb}{1,0.5,0}
\definecolor{violet}{RGB}{70,0,170}
\definecolor{magenta}{RGB}{170,0,170}
\definecolor{dgreen}{RGB}{0,150,0}

\usepackage[normalem]{ulem}

\ifdraft
 \newcommand{\Wei}[1]{{\color{blue}{\bf Wei: #1}}}
 
 \newcommand{\MK}[1]{{\color{magenta}{\bf MK: #1}}}
 \newcommand{\mk}[1]{{\color{magenta} #1}}
\else
 \newcommand{\Wei}[1]{}
 
 \newcommand{\MK}[1]{}
 \newcommand{\mk}[1]{#1}
 
\begin{document}
\title{ActMAD: Activation Matching to Align Distributions for Test-Time-Training}
\author{Muhammad Jehanzeb Mirza$^{1,2}$\and
Pol Jané Soneira$^3$\and
Wei Lin$^{1,4}$\and
Mateusz Kozinski$^1$\and
Horst Possegger$^1$\and
Horst Bischof$^{1,2}$\and\\
{$^1$Institute for Computer Graphics and Vision, TU Graz, Austria.}\\
{$^2$Christian Doppler Laboratory for Embedded Machine Learning.}\\
{$^3$Institute of Control Systems, KIT, Germany.}\\
{$^4$Christian Doppler Laboratory for Semantic 3D Computer Vision.}\\
}
\maketitle
\begin{abstract}
Test-Time-Training (TTT) is an approach to cope with out-of-distribution (OOD) data by adapting a trained model to distribution shifts 
occurring at test-time. 
We propose to perform this adaptation via \emph{Activation Matching} (ActMAD):
We analyze activations of the model and align activation statistics of the OOD test data to those of the training data.
In contrast to existing methods, which model the distribution of entire channels in the ultimate layer of the feature extractor, we model the distribution of each feature in multiple layers across the network.
	This results in a more fine-grained supervision and makes ActMAD attain state of the art performance on CIFAR-100C and Imagenet-C.
ActMAD is also architecture- and task-agnostic, which lets us go beyond image classification, and score 15.4\% improvement over previous approaches when evaluating a KITTI-trained object detector on KITTI-Fog. 
Our experiments highlight that ActMAD can be applied to online adaptation in realistic scenarios, requiring little data to attain its full performance.
\end{abstract}
\blfootnote{Correspondence: \small\tt{muhammad.mirza@icg.tugraz.at}}
\vspace{-0.6cm}
\section{Introduction} \label{sec:intro}

When evaluated in laboratory conditions, deep networks outperform humans in numerous visual recognition tasks~\cite{ecoffet2021first, esteva2017dermatologist, dubey2018investigating}. \MK{Anyone has better citations for this?} But this impressive performance is predicated on the assumption that the test and training data come from the same distribution. In many practical scenarios, satisfying this requirement is either impossible or impractical. For example, a perception system of an autonomous vehicle can be exposed not only to fog, snow, and rain, but also to rare conditions including smoke, sand storms, or substances distributed on the road in consequence of a traffic accident. Unfortunately, distribution shifts between the training and test data can incur a significant performance penalty~\cite{boudiaf2022parameter, liu2021tttpp, mirza2021dua, kojima2022cfa, wang2020tent, wiles2021fine}. To address this shortcoming, test-time-training (TTT) methods attempt to adapt a deep network to the actual distribution of the test data, at test-time.

\begin{figure}
    \centering
    \includegraphics[width=.95\columnwidth]{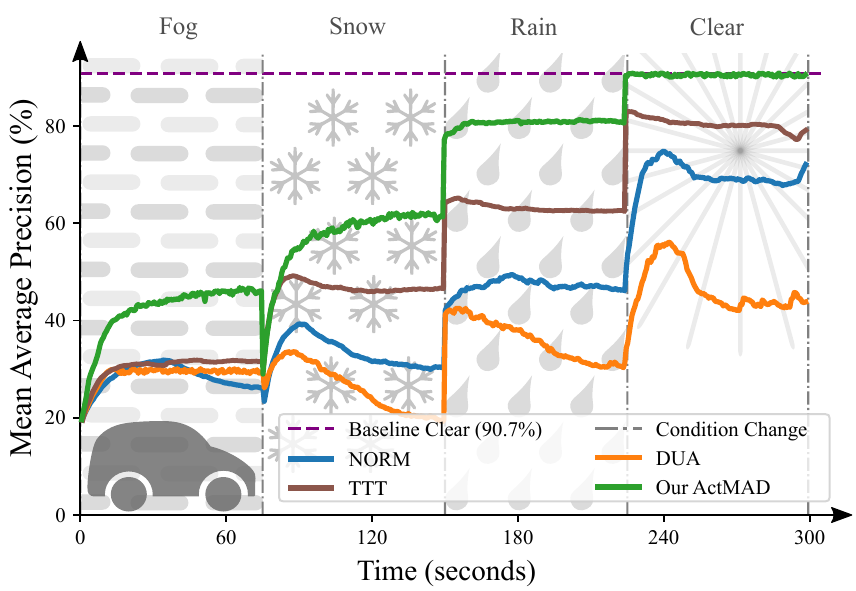}
    \caption{
ActMAD is well suited for online test-time adaptation irrespective of network architecture and task.
Our intended application is object detection on board of a vehicle driving in dynamically changing and unpredictable weather conditions. 
Here, we report the class-averaged Mean Average Precision (mAP@50) over adaptation time\protect\footnotemark, attained by continuously adapting a KITTI~\cite{geiger2013vision}-trained YOLOv3~\cite{redmon2018yolov3} detector. 
\emph{Clear} refers to the weather condition of the original dataset, mostly acquired in sunny weather, while \emph{Fog}, \emph{Rain}, and \emph{Snow} are generated by perturbing the original images~\cite{halder2019physics, hendrycks2019robustness}. Gray vertical lines mark transitions between weather conditions.
    }
    \label{fig:teaser}
\end{figure}

\footnotetext{Time (seconds) for adaptation is hardware specific. For reference, we run these experiments on an RTX-3090 NVIDIA graphics card and it takes $\sim$${75}$s to adapt to $3741$ images (one weather condition) in the KITTI test~set.}

Existing TTT techniques limit the performance drop provoked by the shift in test distribution, but the goal of on-line, data-efficient, source-free, and task-agnostic adaptation remains elusive:
the methods that update network parameters by self-supervision on test data~\cite{sun2020ttt,liu2021tttpp,gandelsman2022test} can only be used if the network was first trained in a multi-task setup,
the approaches that minimise the entropy of test predictions~\cite{liang2020shot,wang2020tent}, or use pseudo-labels, or pseudo-prototypes~\cite{kojima2022cfa,iwasawa2021t3a}, are not easily adapted to object detection or regression, 
and the ones that update statistics of the batch normalization layers~\cite{mirza2021dua,schneider2020norm} are limited to architectures that contain them. 
As a consequence of these limitations, while existing methods fare well in image classification, none of them is well suited for the scenario where the need for on-line adaptation is the most acute, that is, object detection on board a car driving in changing weather conditions.

Our goal is to lift the limitations of the current methods and propose a truly versatile, task-, and architecture-agnostic technique, that would extend TTT beyond image classification and enable its deployment in object detection for automotive applications.
To that end, we revisit feature alignment, the classical domain adaptation technique~\cite{sun2016coral,zellinger2017cmd,bufr}, also adopted by several TTT algorithms~\cite{kojima2022cfa,liu2021tttpp,mirza2021dua,schneider2020norm}. 
It consists in aligning the distribution of test set features to that of the training set. By contrast to previous methods, that align the distributions of entire channels in the ultimate layer of the feature extractor, ActMAD brings the feature alignment approach to another level by individually aligning the distribution of each feature in multiple feature maps across the network.
On \mk{the} one hand, this makes the alignment location-aware. For example, the features from the bottom of the image, most often representing road and other vehicles, are not mixed with features from the top part, more likely to describe trees, buildings, or the sky. On the other hand, it results in a more fine-grained supervision of the adapted network.
Our ablation studies show that ActMAD owes most of its performance to these two contributions.
Additionally, while several authors suggested aligning higher-order moments~\cite{zellinger2017cmd,kojima2022cfa}, we demonstrate that aligning means and variances should be preferred when working with small batches. Unlike methods that only update mean and variance~\cite{mirza2021dua, schneider2020norm}, or affine parameters of batch normalization layers~\cite{mirza2021dua,kojima2022cfa,schneider2020norm}, ActMAD updates all network parameters.
Our approach is architecture- and task-agnostic, and does not require access to the training data or training labels, which is important in privacy-sensitive applications. It does require the knowledge of activation statistics of the training set, but this requirement can be easily satisfied by collecting the data during training, or by computing them on unlabelled data without distribution shift.

Our contribution consists in a new method to use Activation Matching to Align Distributions for TTT, which we abbreviate as ActMAD. Its main technical novelty is that we model the distribution of each point in the feature map, across multiple feature maps in the network.
While most previous works focus on combining different approaches to test-time adaptation, ActMAD is solely based on feature alignment.
This is necessary for our method to be applicable across different architectures and tasks, but we show that discarding the bells and whistles, while aligning activation distributions in a location-aware manner results in matching or surpassing state-of-the-art performance on a number of TTT benchmarks.  
Figure~\ref{fig:teaser} presents the performance of ActMAD in a simulated online adaptation scenario, in which an object detector runs onboard a vehicle driven in changing weather conditions.
Most existing methods cannot be used in this setup, because they cannot run online, or because they cannot be used for object detection. Note, that our method recovers, and even exceeds, the performance of the initial network once the weather cycle goes back to the conditions in which the detector was trained.

\MK{to suppress comments (in bold and color) and remove color from my edits (colored but not bolded), comment out {\textbackslash}drafttrue in includes.tex}
\begin{figure*}
    \centering
    \includegraphics[width=0.85\textwidth]{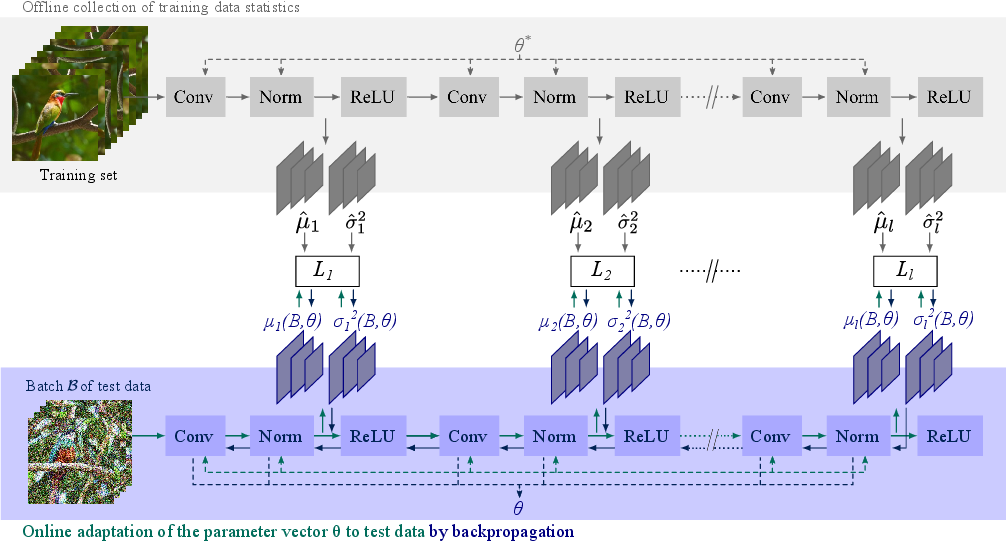}
    \caption{Schematic of ActMAD. Given a pre-trained model and statistics of the clean activations from the training data, it aligns the activation responses from the shifted test data to the clean activations at test-time. We model the activation distributions in terms of the means and variances of each activation, such that the statistics have the same shape as the feature maps. The statistics of the training activations are pre-computed on the training set, or computed on unlabelled data without distribution shift.
    }
    \label{fig:mainfig}
    \vspace{-0.6cm}
\end{figure*}
\section{Related Work}
\label{sec:related}

\paragraph{Unsupervised domain adaptation.}
The idea of aligning neural feature distributions to bridge the domain gap between training and test data was first used for unsupervised domain adaptation (UDA).
In CORAL~\cite{sun2016coral}, the alignment is performed by first whitening the target domain features and then re-coloring them with covariances computed on the source domain. 
Alternative methods train the network to minimize a difference of features statistics, like the higher-order central moments~\cite{zellinger2017cmd}, or the higher-order cumulant tensor~\cite{chen2020homm}.
These UDA algorithms cannot be used for test-time-training (TTT), because they work offline and require access to the source training data and entire test data.
Our ActMAD, and some existing TTT approaches~\eg~\cite{mirza2021dua,liu2021tttpp}, also explicitly compute and align feature statistics of the training and test data, but perform online adaptation without access to the training data.

\paragraph{Test-Time-Training.}
TTT consists in adapting a network to a distribution shift between the test and training data at test-time\MK{\sout{in an online manner} I think we mention in the intro that some of them are not very well suited for online adaptation.} without access to the training data.
The technique that gave the name to the whole group of methods~\cite{sun2020ttt} augments the training routine with a self-supervised task, which enables re-training the network on the test set even though no test labels are available. 
The original work employed rotation prediction~\cite{gidaris2018unsupervised} as the auxiliary task, but subsequent works \cite{gandelsman2022test, liu2021tttpp} replaced it with the Masked Autoencoder reconstruction task~\cite{he2022masked} or contrastive learning~\cite{chen2020simple}. These methods improve performance, but cannot be used if the network was not trained with the auxiliary task. Moreover, contrastive learning requires a large portion of unlabelled data to be effective. 
By contrast, our ActMAD can be used to adapt a network trained with an arbitrary protocol, and requires very little data to attain its full performance.


Foregoing a multi-task training setup, CoTTA~\cite{wang2022cotta} employs a student-teacher setup for test-time adaptation and relies on entropy of the predicted class distribution to transfer the knowledge from teacher to the student. 
ActMAD performs on par with CoTTA on its own benchmark.
However, reliance on entropy makes their setup classification-specific. 

While other methods adapt network parameters, T3A~\cite{iwasawa2021t3a} casts test-time adaptation as prototype learning. 
It relies on pseudo-labels to create test-set-specific pseudo-prototypes, which then substitute the classifier learned on the training set. 
ActMAD outperforms T3A in classifying ImageNet-C and CIFAR-100C images. 
Similar to the authors of the T3A, Boudiaf~\etal~\cite{boudiaf2022parameter} propose LAME, to forego updating the network parameters and develop an alternative prediction strategy 
by generating class likelihoods 
and minimizing an objective function which promotes consistency of predictions coupled with Laplacian regularization. 
LAME is effective for 
alleviating the 
shifts in the prior distribution, but brings little improvement when the shift affects appearance of test images, which is the focus of our work.

Another group of TTT methods adjusts network parameters using entropy-based objectives on the test data. 
SHOT~\cite{liang2020shot} minimizes the expected entropy of individual predictions, while maximizing the entropy of the predicted classes over the entire test set. 
TENT~\cite{wang2020tent} adjusts the scale and shift parameters of batch normalization layers to minimize the entropy of test predictions.
EATA~\cite{eata} proposes to selectively minimize entropy of output predictions. 
ActMAD outperforms SHOT and TENT by a fair margin, but lags behind EATA for image classification, however,
the entropy-based loss functions prevent direct application of these methods to object detection and regression.

For online object detection, MemCLR~\cite{memclr} proposes to use a contrastive learning setup. 
ActMAD out-performs MemCLR in their setup, while being computationally cheaper. 
Interactron~\cite{interactron} is trained to choose actions that let it record optimal new observations.
However, Interactron is not directly comparable with our setup,
because our system cannot take autonomous actions. 

ActMAD is most closely related to approaches that align statistics of training and test features. 
Along these lines, NORM~\cite{schneider2020norm} and DUA~\cite{mirza2021dua} update the batch normalization statistics computed on the training set to match the 
distribution of 
the test set. 
BUFR~\cite{bufr} proposes a bottom-up feature learning strategy to align the features from the source data with those obtained from the distribution shifted test data.
CFA~\cite{kojima2022cfa} updates the affine layers in the network to match the class-conditioned higher-order central moments from the output of the encoder.
The reliance on class-conditioning prevents application of CFA to object detection. 
By contrast, ActMAD is free from this constraint. 
It aligns means and variances of individual features in multiple layers across the network. 
ActMAD outperforms DUA, NORM and BUFR, and matches the performance of CFA in image classification.

\section{ActMAD}
\label{sec:method}

We are given a deep network $f(\vec{x};\vec{\theta})$, where $\vec{x}$ is an input image and $\vec{\theta}$ is a vector of parameters. We are also given a parameter vector $\vec{\theta^*}$, obtained by training the network on a dataset $\datatrain$ of images and their annotations. Our goal is to use $f$ to process a test set $\datatest$ of images that may differ in appearance from the ones in $\datatrain$, but are consistent with them in terms of their content. In many applications, the test images are delivered in a stream, and while we can compose them into batches of moderate size, we are not capable of iterating over the test set.
In our flagship application, $f$ is an object detector deployed in a car, $\datatrain$ is a sequence of images acquired on a sunny day, and $\datatest$ is a video stream acquired during driving the car in changing weather conditions.
To address both the detection and classification scenarios, we impose no constraints on the form of $f(\vec{x};\vec{\theta})$. 
To ensure maximum versatility of ActMAD, we make no assumptions about the architecture of the network, or the training process.

The requirements of task- and architecture-independence rule out the use of techniques that rely on the output of the network to be a probability distribution~\cite{wang2020tent,wang2022cotta,liang2020shot,iwasawa2021t3a}, or employ auxiliary tasks during training~\cite{liu2021tttpp,sun2020ttt}.
We thus follow the only viable design choice: we treat selected layers of the network as random variables and align the distribution of these variables in the test set to that of the training set.
An overview of our method is presented in Figure.~\ref{fig:mainfig}. 

\paragraph{Location-aware activation alignment.}
We denote the activation responses of the ${l\text{-th}}$ layer of the network $f$ computed for image $\vec{x}$ by $a_l (\vec{x};\vec{\theta})$.
This is a feature map of shape ${C \times W \times H}$, denoting the number of channels and its spatial dimensions.
Typically, $a_l$ does not depend on all model parameters $\vec{\theta}$, but we abuse the notation in the interest of simplicity.
Existing TTT methods based on feature alignment~\cite{mirza2021dua,schneider2020norm,liu2021tttpp,kojima2022cfa} treat all the features in a channel as instantiations of the same random variable, irrespective of their location in the image.
By contrast, we assume that different locations in the feature map may be distributed differently.
This is particularly the case in photographs, because humans frame objects of interest when taking pictures, and in driving scenarios, where different objects occur in different regions of the image.
We therefore separately align each of the $C \times W \times H$ activations in the ${l\text{-th}}$ layer. This results in a stronger supervisory signal than the classical approach of integrating over the spatial dimensions and aligning the distributions of $C$ channels at the output of the encoder. Our experiments show that this stronger supervision leads to faster and more effective adaptation.

\paragraph{Aligning means and variances.}
To align activations, we could use one of the existing feature alignment methods, for example adversarial alignment~\cite{ganin2016domain}, but this would require access to the training data, which we are not granted. We are therefore limited to methods that characterize the distribution in terms of statistics, which could be pre-computed on the training data, and enforce distribution alignment by minimizing the discrepancy of the test statistics with those of the training set. 
The classical way to do this is to minimize the difference between higher-order sample moments of the distributions~\cite{zellinger2017cmd,chen2020homm}.
However, unlike in domain adaptation, where moment-based methods gained popularity, we need to adapt the distributions based on small samples, since we model each location in each channel separately, and run the adaptation online, on small batches.
We thus forego higher-order statistics and use feature means 
\begin{equation}
\mu_l(\dataset;\vec{\theta})=\frac{1}{N_\dataset} \sum_{\vec{x}\in \dataset} a_l(\vec{x};\vec{\theta})
\end{equation}
and variances
\begin{equation} 
\sigma^2_l(\dataset;\vec{\theta})=\frac{1}{N_\dataset} \sum_{\vec{x}\in \dataset} \Bigl(a_l(\vec{x};\vec{\theta})-\mu_l(\dataset;\vec{\theta})\Bigr)^2 ,
\end{equation} 
where $N_\dataset$ is the number of images in $\dataset$, \ie~the dataset over which the estimates are computed.
For each selected layer $l$, we pre-compute the activation statistics of the training set $\datatrain$,
\begin{equation}
\trainmean = \mu(\datatrain;\vec{\theta^*}) \qquad\text{and}\qquad \trainvar = \sigma^2_l(\datatrain;\vec{\theta^*}).
\end{equation}
During adaptation, we compute the same statistics for each batch $\databatch$ of the test set, and minimize 
\begin{equation} \label{eq:layer_objective}
L_l(\databatch;\vec{\theta})=\left| \mu_l(\databatch;\vec{\theta}) - \trainmean \right| + \left| \sigma^2_l(\databatch;\vec{\theta})-\trainvar \right|,
\end{equation}
where $|\cdot|$ denotes the element-wise L1 norm.
\begin{table*}
\setlength\tabcolsep{3.5pt} 
\centering
\small
\begin{tabular}{@{}l@{\hspace{2em}}ccccccccccccccc|c@{\hspace{0.1\tabcolsep}}l}
      \multicolumn{1}{r}{Corruptions:}            &  Gauss &  Shot  &  Impul &  Defcs &   Gls  &   Mtn  &   Zm   &   Snw  &  Frst  &   Fg   &  Brt   &   Cnt  &   Els  &   Px   &   Jpg  &  Mean \\
      \midrule
      Source      & {28.8} & {22.9} & {26.2} & { 9.5} & {20.6} & {10.6} & { 9.3} & {14.2} & {15.3} & {17.5} & { 7.6} & {20.9} & {14.7} & {41.3} & {14.7} & {18.3}\\
      \midrule
      \makebox[\widthof{TTT++}][l]{SHOT$^{\dagger}$} (Offline)  & {13.4} & {11.6} & {16.3} & { 7.3} & {15.9} & { 8.2} & { 7.1} & { 9.4} & { 9.4} & {10.2} & { 6.3} & { 8.3} & {12.8} & { 9.8} & {13.6} & {10.6}\\
      TTT++$^{\dagger}$ (Offline)  & {12.8} & {11.1} & {11.2} & { 7.3} & {17.1} & { 8.2} & { 6.5} & { 9.4} & { 9.9} & { 7.9} & { 5.0} & { 5.1} & {13.7} & { 8.8} & {10.6} & { 9.6}\\
\midrule
      DUA$^*$ &             {15.4} &             {13.4} & {\underline{17.3}} &             {8.0} &             {18.0} &             {9.1} & {\underline{7.7}} &             {10.8} &            {10.8} &             {12.1} &             {6.6} &            {10.9} &             {13.6} &             {13.0} &            {14.3} &             {12.1} &$\pm$0.1\\
   NORM$^*$ &             {15.9} &             {13.7} &             {18.0} &             {7.8} &             {18.3} &             {8.9} &             {8.0} &             {10.8} & {\underline{9.6}} &             {12.7} & {\underline{6.1}} &             {9.4} &             {13.5} &             {14.3} &            {14.5} &             {12.1}&$\pm$0.01 \\
    T3A$^*$ &             {15.7} &             {13.9} &             {17.8} &             {7.9} &             {18.2} &             {9.0} &             {8.2} &             {10.9} &             {9.7} &             {12.6} & {\underline{6.1}} &             {9.2} & {\underline{13.4}} &             {14.2} &            {14.4} &             {12.1}&$\pm$0.4 \\
    P-L$^*$ &             {15.8} &             {14.1} &             {17.8} &             {7.8} &             {18.1} &             {8.9} &             {8.0} &             {10.8} &             {9.7} &             {12.4} & {\underline{6.1}} &             {9.3} & {\underline{13.4}} &             {14.1} &            {14.5} &             {12.0}&$\pm$0.2 \\
    CFA$^{\dagger}$ &             {15.8} &             {13.8} &             {17.9} &             {7.8} &             {18.2} &             {9.0} &             {8.1} &             {10.7} & {\underline{9.6}} &             {12.4} & {\underline{6.1}} &             {9.3} &             {13.5} &             {13.7} &            {14.5} &             {12.0}&$\pm$0.01 \\
   BUFR$^{\dagger}$ &             {18.5} &             {16.3} &             {22.6} &             {9.0} &             {21.8} &            {10.4} &             {9.7} &             {12.7} &            {13.4} &             {15.2} &             {7.5} &            {12.0} &             {16.3} &             {15.1} &            {17.5} &             {14.5}&$\pm$0.01 \\
  TTT++$^{\dagger}$ (Online) &             {15.5} &             {14.1} &             {23.6} &             {9.1} &             {25.1} &            {11.4} &             {8.1} &             {13.2} &            {13.1} &             {13.4} &             {6.6} &  {\bfseries{6.9}} &             {17.6} &             {12.5} & {\bfseries{13.6}} &             {13.6}&$\pm$0.03 \\
SHOT$^{\dagger}$ (Online) & {\underline{14.5}} & {\underline{12.3}} &             {17.7} &             {7.8} &             {17.8} & {\underline{8.7}} &             {7.9} &             {10.4} & {\underline{9.6}} &             {12.1} & {\underline{6.1}} &             {9.0} & {\underline{13.4}} &             {11.4} &            {14.4} & {\underline{11.5}}&$\pm$0.02 \\
   TENT$^*$ & {\underline{14.5}} &             {12.4} &             {17.7} & {\underline{7.7}} & {\underline{17.7}} &             {8.8} &             {7.9} & {\underline{10.3}} & {\underline{9.6}} & {\underline{12.0}} & {\underline{6.1}} &             {9.0} & {\underline{13.4}} & {\underline{11.3}} &            {14.5} & {\underline{11.5}}&$\pm$0.02 \\
      ActMAD$^{\dagger}$ &  {\bfseries{13.0}} &  {\bfseries{11.2}} &  {\bfseries{15.1}} &  {\bfseries{7.4}} &  {\bfseries{15.9}} &  {\bfseries{8.3}} &  {\bfseries{7.1}} &   {\bfseries{9.5}} &  {\bfseries{9.3}} &  {\bfseries{10.6}} &  {\bfseries{5.9}} & {\underline{8.4}} &  {\bfseries{12.3}} &   {\bfseries{9.3}} & {\bfseries{13.6}} &  {\bfseries{10.4}}&$\pm$0.06 \\
      \bottomrule
\end{tabular}
\caption{Top-1 Classification Error (\%) for all corruptions in CIFAR-10C (level 5). Lower is better. All results are for a WRN-40-2 backbone. \emph{Source} denotes the performance on the corrupted test data without any adaptation. 
For ease of readability, we highlight the lowest error in bold and the second best as underlined.} 
\label{tab:cifar-10C-results} 
\vspace{-0.3cm}
\end{table*}



\begin{table*}
\setlength\tabcolsep{3.5pt}
\centering
\small
\begin{tabular}{@{}l @{\hspace{2em}}ccccccccccccccc |c@{\hspace{0.1\tabcolsep}}l}
      \multicolumn{1}{r}{Corruptions:}            &  Gauss &  Shot  &  Impul &  Defcs &   Gls  &   Mtn  &   Zm   &   Snw  &  Frst  &   Fg   &  Brt   &   Cnt  &   Els  &   Px   &   Jpg  &  Mean \\
      \midrule
      Source &             {65.7} &             {60.1} &             {59.1} &            {32.0} &             {51.0} &             {33.6} &             {32.4} &             {41.4} &             {45.2} &             {51.4} &             {31.6} &             {55.5} &             {40.3} &             {59.7} &             {42.4} &             {46.7}  & \\
      \midrule
      \makebox[\widthof{TTT++}][l]{SHOT$^{\dagger}$} (Offline)  & {37.2} & {36.2} & {36.7} & {27.5} & {38.2} & {28.5} & {27.8} & {31.8} & {32.0} & {33.4} & {25.8} & {29.6} & {34.5} & {29.8} & {37.2} & {32.4} & \\
 TTT++$^{\dagger}$ (Offline) & {40.7} & {36.4} & {41.5} & {27.5} & {47.8} & {31.1} & {25.1} & {36.5} & {34.7} & {33.7} & {23.3} & {24.7} & {40.2} & {30.5} & {33.3} & {33.8} & \\
      \midrule

         DUA$^*$  &             {42.2} &             {40.9} & {\underline{41.0}} &            {30.5} &             {44.8} &             {32.2} &             {29.9} &             {38.9} &             {37.2} &            {43.6} &             {29.5} &             {39.2} &             {39.0} &             {35.3} &             {41.2} &             {37.7} &$\pm$0.3\\
   NORM$^*$  &             {42.5} &             {41.8} &             {42.6} &            {29.7} &             {43.9} &             {30.6} &             {29.7} &             {35.7} &             {34.6} &            {42.2} &             {26.9} &             {32.8} &             {38.1} &             {35.5} &             {40.9} &             {36.5}&$\pm$0.03 \\
    T3A$^*$ &             {42.4} &             {41.8} &             {42.5} &            {29.7} &             {44.3} &             {30.5} &             {29.5} &             {35.9} &             {34.5} &            {42.1} &             {26.8} &             {32.8} &             {38.0} &             {35.9} &             {40.7} &             {36.5}&$\pm$0.02 \\
    P-L$^*$  &             {41.3} &             {40.5} &             {42.5} &            {29.6} &             {43.1} &             {30.3} &             {29.4} &             {35.8} &             {34.3} &            {41.7} & {\underline{26.7}} &             {32.4} &             {37.8} &             {33.5} &             {40.8} &             {36.0}&$\pm$0.05 \\
    CFA$^{\dagger}$ &             {40.4} &             {39.3} &             {42.1} &            {29.4} &             {42.3} & {\underline{30.2}} & {\underline{29.2}} &             {35.1} & {\underline{34.1}} &            \underline{39.8} & {\underline{26.7}} &             {32.1} &             {37.6} &             {32.8} &             {40.6} &             {35.5} &$\pm$0.01\\
   BUFR$^{\dagger}$ &             {44.5} &             {44.3} &             {47.5} &            {32.4} &             {45.9} &             {33.0} &             {33.1} &             {38.7} &             {38.2} &            {45.7} &             {30.1} &             {36.1} &             {40.7} &             {37.1} &             {44.3} &             {39.4}&$\pm$0.03 \\
  TTT++$^{\dagger}$ (Online) &             {43.9} &             {40.0} &             {56.3} &            {32.5} &             {54.2} &             {35.9} &             {29.9} &             {42.2} &             {39.4} & {\bfseries{39.7}} &             {27.5} &  {\bfseries{29.6}} &             {44.2} &             {37.0} &  {\bfseries{37.4}} &             {39.3} &$\pm$0.04\\
SHOT$^{\dagger}$ (Online) & {\underline{39.7}} & {\underline{38.9}} &             {42.1} & {\bfseries{29.0}} & {\underline{41.9}} & {\underline{30.2}} &             {29.3} & {\underline{34.8}} &             {34.2} & {\bfseries{39.7}} & {\underline{26.7}} &             {32.2} & {\underline{37.2}} &             {32.5} &             {40.4} & {\underline{35.3}} &$\pm$0.1\\
   TENT$^*$ &             {39.9} &             {39.1} &             {42.2} & {\bfseries{29.0}} &             {42.0} & {\underline{30.2}} &             {29.3} &             {34.9} &             {34.2} & {\bfseries{39.7}} & {\underline{26.7}} &             {32.3} &             {37.4} & {\underline{32.4}} &             {40.4} & {\underline{35.3}}&$\pm$0.03\\
 ActMAD$^{\dagger}$ &  {\bfseries{39.6}} &  {\bfseries{38.4}} &  {\bfseries{39.5}} &            \underline{29.1} &  {\bfseries{41.5}} &  {\bfseries{30.0}} &  {\bfseries{29.1}} &  {\bfseries{34.0}} &  {\bfseries{33.2}} &            {40.2} &  {\bfseries{26.4}} & {\underline{31.5}} &  {\bfseries{36.4}} &  {\bfseries{31.4}} & {\underline{38.9}} &  {\bfseries{34.6}} &$\pm$0.01 \\
\bottomrule
\end{tabular}
\caption{Top-1 Classification Error (\%) for all corruptions in CIFAR-100C (level 5). Lower is better. The results were obtained by adapting a WRN-40-2 backbone, trained on CIFAR100, to CIFAR-100C.}
\label{tab:cifar-100C-results}
\vspace{-0.2cm}
\end{table*}

\begin{table}
\setlength\tabcolsep{1.0pt}
\centering
\small
\begin{tabular}{lccccccc}
      \multicolumn{1}{r}{Methods:} &Source&TENT$^*$&SHOT$^{\dagger}$&CFA$^{\dagger}$&ActMAD$^{\dagger}$\\
      \toprule

      CIFAR-10C&14.6$\pm$0.2&10.9$\pm$0.2&8.9$\pm$0.1&\underline{8.4}$\pm$0.01&\textbf{7.7}$\pm$0.1\\
      \midrule
      CIFAR-100C&35.1$\pm$0.3&27.4$\pm$0.2&25.6$\pm$0.1&\underline{24.6}$\pm$0.03&\textbf{22.4}$\pm$0.1\\
      \bottomrule
\end{tabular}
\caption{Mean Top-1 Classification Error (\%) over all corruptions in CIFAR-10/100C (level 5). Lower is better. All results are for a ViT-B/16 backbone. 
ActMAD results are averaged over 10 runs. 
}
\label{tab:cifar-10-100C-results-vit}

\end{table}
\paragraph{Aligning multiple layers.}
We found that aligning just the statistics of the final layer, following the common practice~\cite{liu2021tttpp,kojima2022cfa}, is not enough for fast and effective distribution alignment. We show this in an ablation study in Section~\ref{sec:ablations}.
We therefore define a set $\mathcal{L}$ of multiple layers for alignment and the corresponding alignment objective as
\begin{equation}
L(\databatch;\vec{\theta})=\sum_{l \in \mathcal{L}} L_l(\databatch;\vec{\theta}) .
\end{equation}
Intuitively, the aligned layers should be distributed evenly across the network. In the case of convolutional architectures with the standard convolution--normalization--ReLU sequences, we found that it is best to align the features obtained after normalization and before the nonlinearities.
\paragraph{Updating the complete parameter set.}
Finally, while numerous existing methods only update affine parameters or statistics of the batch normalization layers~\cite{kojima2022cfa,schneider2020norm,mirza2021dua,wang2020tent}, we update all the parameters of the adapted network.
It might seem \emph{a priori} that this increases the risk of making the parameter vector drift away from its initial value, but continuously aligning a very rich set of statistics actually allows the network to recover its initial performance after the distribution of the test data becomes consistent with that of the training set, as shown in Figure.~\ref{fig:teaser}.
The ablation studies in Section~\ref{sec:ablations} further confirm that there is no significant performance difference between the two strategies, and
thus, we opt for the more deployment-friendly of the two alternatives and make a gradient step on
\begin{equation}
\min_\vec{\theta} L(\databatch;\vec{\theta})
\end{equation}
for each batch $\databatch$ of the test data $\datatest$.

\begin{table}
\setlength\tabcolsep{1.5pt}
\centering
\small
\begin{tabular}{@{}l @{\hspace{1em}}ccccccc
@{\hspace{1.0em}}c@{}c@{}}
      \multicolumn{1}{r}{Methods:} &Source&NORM$^*$&TENT$^*$&CoTTA$^*$&ActMAD$^\dagger$\\
      \toprule
      CIFAR-10C&18.3&13.8$\pm$0.4&12.1$\pm$0.7&\underline{11.5}$\pm$0.1&\textbf{11.4}$\pm$0.1\\
      \midrule
      CIFAR-100C&46.7&42.6$\pm$0.8&38.2$\pm$0.7&\textbf{35.7}$\pm$0.4&\underline{35.9}$\pm$0.5\\
      \bottomrule
\end{tabular}
\caption{Continuous Adaptation (following CoTTA~\cite{wang2022cotta}) to distribution shifts at test-time. \mk{We report the mean error of 10 runs, while shuffling the order of the corruptions for each run.}
}
\label{tab:cifar-10-100C-results-cont-adapt}
\vspace{-0.3cm}
\end{table}

\begin{table*}
\setlength\tabcolsep{3.5pt}
\centering
    \small
\begin{tabular}{@{}l@{\hspace{2em}}ccccccccccccccc|c@{\hspace{0.1\tabcolsep}}l}
      \multicolumn{1}{r}{Corruptions:}            &  Gauss &  Shot  &  Impul &  Defcs &   Gls  &   Mtn  &   Zm   &   Snw  &  Frst  &   Fg   &  Brt   &   Cnt  &   Els  &   Px   &   Jpg  &  Mean \\
      \midrule
      Source &             {98.4} &             {97.7} &             {98.4} &             {90.6} &             {92.5} &             {89.8} &             {81.8} &             {89.5} &             {85.0} &             {86.3} &             {51.1} &             {97.2} &             {85.3} &             {76.9} &             {71.7} &             {86.2} & \\

\midrule
SHOT$^\dagger$ (Offline) & {73.8} & {70.5} & {72.2} & {79.2} & {80.6} & {58.5} & {54.0} & {53.6} & {63.0} & {47.3} & {39.2} & {97.7} & {48.7} & {46.1} & {53.0} & {62.5} & \\
\midrule
         TTT$^\dagger$ &             {96.9} &             {95.5} &             {96.5} &             {89.9} &             {93.2} &             {86.5} &             {81.5} &             {82.9} &             {82.1} &             {80.0} &             {53.0} &             {85.6} &             {79.1} &             {77.2} &             {74.7} &             {83.6} & \\

         DUA$^*$ &             {89.4} &             {87.6} &             {88.1} &             {88.0} &             {88.6} &             {84.7} &             {74.3} &             {77.8} &             {78.4} &             {68.6} &             {45.6} &             {95.9} &             {72.2} &             {66.5} &             {67.4} &             {78.2} & $\pm$0.7 \\
        NORM$^*$ &             {87.1} &             {90.6} &             {89.5} &             {87.6} &             {93.4} &             {80.0} &             {71.9} &             {70.6} &             {81.5} &             {65.9} &             {46.8} &             {89.8} &             {73.5} &             {63.2} &             {67.5} &             {77.3}  & $\pm$0.3 \\
         T3A$^*$ &             {85.5} &             {84.0} &             {85.0} &             {86.6} &             {85.9} &             {76.1} &             {65.4} &             {70.3} &             {71.0} &             {58.7} &             {41.3} &             {86.8} &             {60.5} &             {54.4} &             {61.0} &             {71.5}  & $\pm$0.5 \\
     SHOT$^\dagger$  (Online) &             {83.9} &             {82.3} &             {83.7} &             {83.9} &             {83.8} &             {72.6} &             {61.9} &             {65.7} &             {68.6} &             {54.8} & {\underline{39.4}} &             {85.9} &             {58.1} &             {53.1} &             {62.3} &             {69.3}  & $\pm$0.03\\
P-L$^*$  &             {82.0} &             {79.7} &             {81.5} &             {84.2} &             {83.0} & {\underline{71.0}} &             {60.7} &             {65.4} &             {68.6} &             {52.9} &             {41.7} &             {82.6} & {\underline{55.5}} &             {51.1} & {\underline{55.7}} &             {67.7}  & $\pm$0.3\\
        TENT$^*$  &             {80.8} &             {78.6} &             {80.4} &             {82.5} &             {82.5} &             {72.1} & {\underline{60.5}} & {\underline{63.7}} & {\underline{66.7}} & {\underline{52.1}} &  {\bfseries{39.2}} &             {84.2} & {\underline{55.5}} & {\underline{50.8}} &             {58.2} &             {67.2}  & $\pm$0.02\\
         CFA$^\dagger$  & {\underline{78.2}} &  {\bfseries{76.4}} & {\underline{78.2}} & {\underline{81.9}} & {\underline{80.4}} &  {\bfseries{69.6}} &  {\bfseries{60.1}} &  {\bfseries{63.4}} &             {67.6} &  {\bfseries{52.0}} &             {41.5} & {\underline{79.5}} &  {\bfseries{54.3}} &  {\bfseries{50.2}} &  {\bfseries{55.1}} &  {\bfseries{65.9}}  & $\pm$0.01 \\
      ActMAD$^\dagger$  &  {\bfseries{76.3}} & {\underline{77.4}} &  {\bfseries{77.4}} &  {\bfseries{76.1}} &  {\bfseries{75.4}} &             {72.0} &             {62.8} &             {66.6} &  {\bfseries{65.8}} &             {55.8} &             {40.9} &  {\bfseries{78.8}} &             {55.7} &             {51.4} &             {57.6} & {\underline{66.0}} & $\pm$0.02 \\
\midrule
\end{tabular}
\caption{Top-1 Classification Error (\%) for all corruptions in ImageNet-C (level 5). Lower is better. All results are for a ResNet-18 network pre-trained on the clean train set. \emph{Source} denotes its performance on the corrupted test data without any adaptation.}
\label{tab:imagenet-c-results}
\vspace{-0.4cm}
\end{table*}

\section{Experimental Evaluation}
\label{sec:results}
We first summarize the datasets and baselines used for the evaluation, and then discuss our results.
\renewcommand*{\thefootnote}{\fnsymbol{footnote}} 
\footnotetext[1]{Fully Test-Time Adaptation approaches} 
\footnotetext[2]{Approaches requiring some supervision from the training data}

\subsection{Datasets}

\noindent
We ran experiments on the following data sets:
\begin{itemize}[nosep,label=-,wide]
\item \emph{CIFAR-10C and CIFAR-100C}\cite{hendrycks2019robustness} were created by corrupting test images of the CIFAR-10 and CIFAR-100 datasets~\cite{krizhevsky2009learning}. We used them to benchmark the performance of ActMAD in image classification, and for ablation studies. Following the common protocol~\cite{mirza2021dua, wang2020tent, sun2020ttt, liu2021tttpp, schneider2020norm}, we used 15 corruption types and 5 corruption levels.
\item \emph{ImageNet-C}~\cite{hendrycks2019robustness} contains ImageNet~\cite{deng2009imagenet} test images, corrupted in the same way as those of CIFAR-10/100C. We used them for large-scale classification experiments. Again, we tested for 15 corruption types and 5 severity levels. 
\item We used the  \emph{KITTI} dataset~\cite{geiger2013vision} to evaluate the performance of ActMAD on object detection for automotive applications. KITTI contains 8 classes of traffic participants, 
and was recorded mostly in clear weather. To simulate degrading weather conditions on the KITTI dataset, we used physics-based rendering of fog and rain~\cite{halder2019physics}, and followed Mirza et al.~\cite{mirza2022incremental} to simulate snow.
\end{itemize}

 For KITTI, the annotations for the test set are not publicly available, so 
 we divided the public train set randomly into train and test split with equal proportion. 
 These splits and our codebase is available at this repository: \href{https://github.com/jmiemirza/ActMAD}{https://github.com/jmiemirza/ActMAD}.

\subsection{Baselines}
\noindent
We compared ActMAD against the following approaches:
\begin{itemize}[nosep,label=-]
\item \emph{Source} is the model trained on source data and evaluated on test data without any adaptation. 
\item \emph{TTT}~\cite{sun2020ttt} adapts the network to test data using self-supervision from a rotation prediction task. 
\item \emph{TTT++}~\cite{liu2021tttpp} employs a combination of self-supervised contrastive learning and feature alignment. 
\item \emph{P-L} is our adaptation of the pseudo-labeling approach~\cite{lee2013pseudo} to test-time training. 
\item \emph{CFA}~\cite{kojima2022cfa} aligns class-conditioned feature statistics. 
\item \emph{BUFR}~\cite{bufr} relies on aligning a lightweight approximation of train and test data statistics.  
\item \emph{T3A}~\cite{iwasawa2021t3a} uses pseudo-prototypes to classify test data. 
\item \emph{SHOT}~\cite{liang2020shot} minimizes prediction entropy while maximizing class entropy over the test data.
\item \emph{NORM}~\cite{schneider2020norm} adapts the statistics of batch normalization layers to test data.
\item \emph{DUA}~\cite{mirza2021dua} is another approach to adapt the batch normalization statistics online.
\item \emph{TENT}~\cite{wang2020tent} learns channel-wise affine transformations by minimizing the entropy of the predictions.
\end{itemize}
ActMAD experiments are run for $10$ random seeds, while the online baselines are run for $3$ random seeds (except for TTT~\cite{sun2020ttt} for 
ImageNet~\cite{deng2009imagenet}, since it requires retraining).
We report the mean and standard deviations for these runs.  

\subsection{Image Classification}

To evaluate ActMAD on image classification, we adapted a Wide-ResNet-40-2~\cite{zagoruyko2016wide} (WRN-40-2) from the Robust Bench~\cite{hendrycks2019robustness} and a Vision Transformer~\cite{dosovitskiy2020image} (ViT-B/16) backbone on the CIFAR-C datasets.
\MK{Big Horst is asking where he can see the performance difference between VIT and ResNet. This is a pre-cautious comment: did you include all the results. I think that the best way to address it would be to embed the results from Tab3 into Tab1 and Tab2. This would make the results more readable than now, IMO. It would also necessitate moving Tab1 into the appendix, due to space limitations. If you do not want to do this, I suggest you ignore this comment. The Tabs with transformer and resnet results are specified in the text below.}
For Imagenet-C dataset, we used an ImageNet pre-trained ResNet-18~\cite{he2016deep} from the PyTorch model zoo, following previous work~\cite{mirza2021dua}.
Results on ImageNet-C with other backbones are provided in the supplemental. 
We used a batch size of $128$ for all image classification experiments. 
ActMAD was run with the learning rate of $1\mathrm{e}{-2}$ for CIFAR-10C, $1\mathrm{e}{-3}$ for CIFAR-100C and $25\mathrm{e}{-5}$ for ImageNet-C, while the baselines used hyper-parameters reported by their authors. 
Following prior literature~\cite{mirza2021dua, wang2020tent, iwasawa2021t3a}, we performed the experiments online, using each test image to adapt the network only once.
Since SHOT and TTT++ are better suited for the offline scenario, we additionally run them for 500 epochs on the entire test sets of the CIFAR-C data sets, with an early stopping criterion, according to the official TTT++ implementation\footnote{\href{https://github.com/vita-epfl/ttt-plus-plus/tree/e429a194}{https://github.com/vita-epfl/ttt-plus-plus}, commit: e429a194}. 

\begin{figure*}
\begin{minipage}{0.55\textwidth}
\setlength\tabcolsep{3.7pt}
\centering
\small
\begin{tabular}{@{}l cccccccc|c@{\hspace{0.1\tabcolsep}}l}
\multicolumn{11}{c}{(a) KITTI-Clear $\to$ KITTI-Fog}\\
       &                car &                van &              truck &         ped &     sit &            cyc &               tram &               misc &               Mean & \\
\midrule
Source &             {31.3} &             {15.0} &              {6.0} &             {34.8} &             {33.6} &             {20.2} &              {6.7} &              {9.1} &             {19.6} \\
   TTT$^\dagger$ &             {42.6} &             {19.5} &             {10.5} & {\underline{49.7}} &             {51.4} &             {31.0} &             {10.5} &             {20.2} &             {29.4}&$\pm$0.5 \\
   DUA$^*$ & {\underline{51.4}} &             {13.5} &              {9.1} &             {48.1} & {\underline{57.3}} & {\underline{36.3}} &             {14.5} &             {18.0} &             {31.0} &$\pm$2.1\\
  NORM$^*$ &             {50.1} & {\underline{27.6}} & {\underline{12.6}} &             {47.6} &             {50.0} &             {30.9} & {\underline{17.7}} & {\underline{21.7}} & {\underline{32.3}}&$\pm$0.1 \\
ActMAD$^\dagger$ &  {\bfseries{67.0}} &  {\bfseries{41.2}} &  {\bfseries{25.5}} &  {\bfseries{62.2}} &  {\bfseries{68.7}} &  {\bfseries{50.9}} &  {\bfseries{30.5}} &  {\bfseries{35.7}} &  {\bfseries{47.7}}&$\pm$1.01 \\
\midrule
\\[-0.8em]

\multicolumn{11}{c}{(b) KITTI-Clear $\to$ KITTI-Rain}\\
\midrule
Source &             {86.4} &             {69.6} &             {58.6} &             {68.6} &             {63.7} &             {60.2} &             {64.5} &             {60.4} &             {66.5} \\
   TTT$^\dagger$ &             {86.4} &             {76.1} &             {68.0} &             {68.7} & {\underline{66.6}} &             {66.3} & {\underline{75.0}} &             {65.2} &             {71.5}&$\pm$0.4 \\
   DUA$^*$ & {\underline{88.3}} &             {70.4} & {\underline{70.4}} & {\underline{70.8}} &  {\bfseries{67.7}} & {\underline{66.8}} &             {73.5} & {\underline{67.5}} & {\underline{71.9}}&$\pm1.3$ \\
  NORM$^*$ & {\underline{88.3}} & {\underline{77.0}} &             {65.7} &             {69.1} &             {61.5} &             {66.7} &             {69.1} &             {67.1} &             {70.6}&$\pm0.1$  \\
ActMAD$^\dagger$ &  {\bfseries{94.2}} &  {\bfseries{89.2}} &  {\bfseries{87.3}} &  {\bfseries{74.1}} &             {65.6} &  {\bfseries{77.9}} &  {\bfseries{82.5}} &  {\bfseries{80.1}} &  {\bfseries{81.4}}&$\pm0.2$\\
\midrule
\\[-0.8em]

\multicolumn{11}{c}{(c) KITTI-Clear $\to$ KITTI-Snow}\\
\midrule
Source &             {54.8} &             {27.8} &             {31.7} &             {35.7} &              {1.3} &             {15.7} &             {18.2} &             {13.3} &             {24.8} \\
   TTT$^\dagger$ & {\underline{77.2}} & {\underline{53.2}} & {\underline{60.6}} & {\underline{48.4}} & {\underline{29.7}} & {\underline{37.1}} &             {43.2} &             {31.1} & {\underline{47.5}}&$\pm$0.7 \\
   DUA$^*$ &             {64.6} &             {38.9} &             {49.3} &             {44.0} &             {20.8} &             {22.8} &             {27.8} &             {25.4} &             {36.7} & $\pm0.9$\\
  NORM$^*$ &             {75.5} &             {51.0} &             {51.7} &             {46.8} &             {21.7} &             {34.9} & {\underline{43.4}} & {\underline{34.3}} &             {44.9}& $\pm0.2$ \\
ActMAD$^\dagger$ &  {\bfseries{89.5}} &  {\bfseries{78.0}} &  {\bfseries{82.6}} &  {\bfseries{57.8}} &  {\bfseries{38.0}} &  {\bfseries{53.4}} &  {\bfseries{58.8}} &  {\bfseries{58.1}} &  {\bfseries{64.5}}& $\pm0.03$ \\
\midrule
\end{tabular}
\captionsetup{type=table}
\caption{Mean Average Precision (mAP@50) for a KITTI pre-trained YOLOv3 tested on rain, fog and snow datasets. Higher is better. a)~Results for the most severe fog level,~\ie only $30$m visibility. b) Results for the most severe rain level,~\ie $200$mm/hr rain intensity. c) Results for snow. Best mAP is shown in bold, while the second best is underlined.
}
\label{tab:kitti_fog_rain_snow}
\end{minipage}
    \hfill
\begin{minipage}{0.40\textwidth}
\vspace{-0.55cm}
        \includegraphics[scale=0.35,trim = 0 15 0 0, clip]{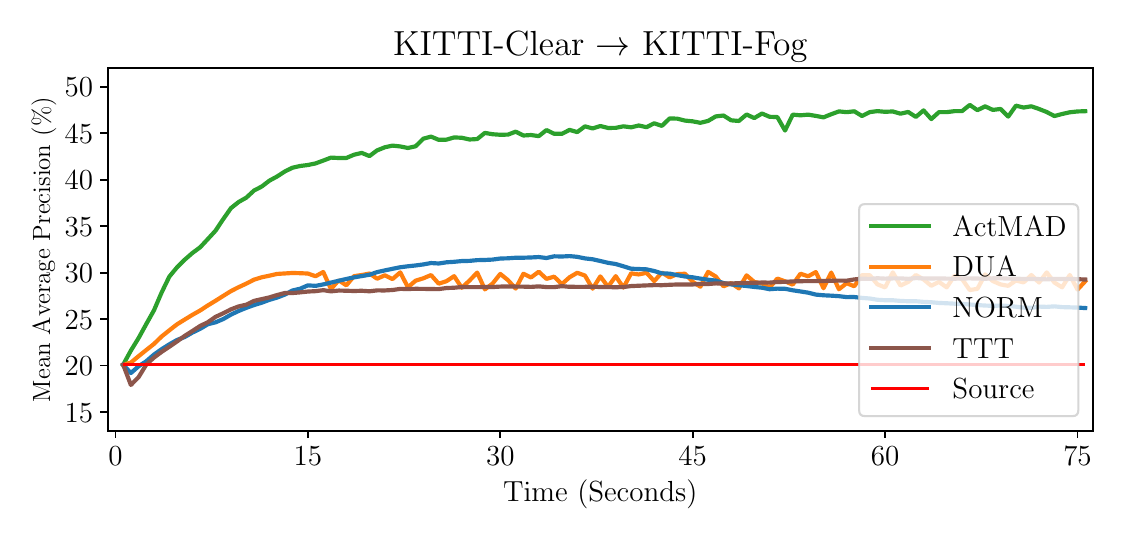}

        \includegraphics[scale=0.35,trim = 0 15 0 0, clip]{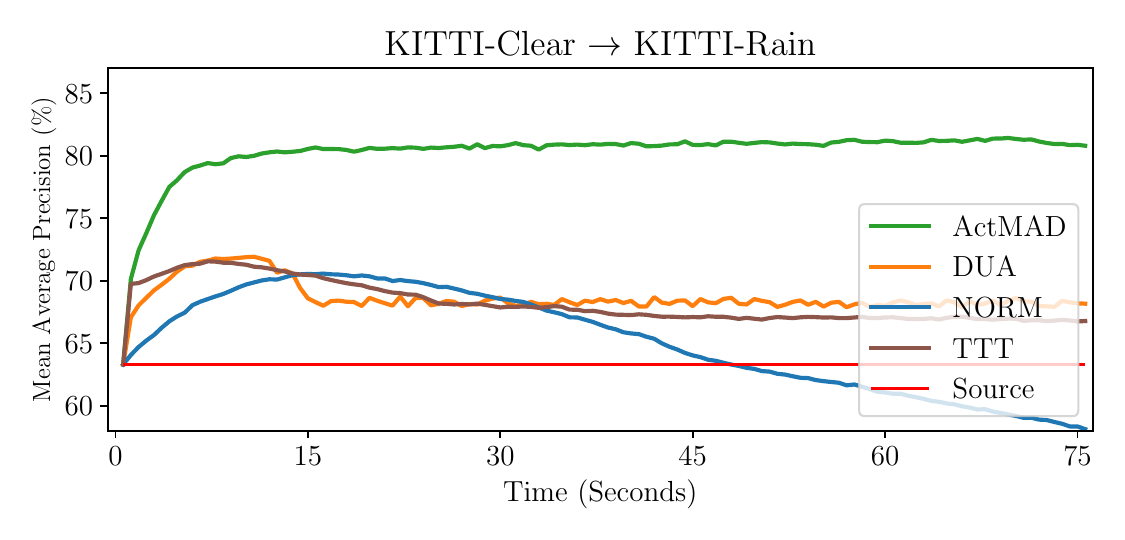}

        \includegraphics[scale=0.35,trim = 0 15 0 0, clip]{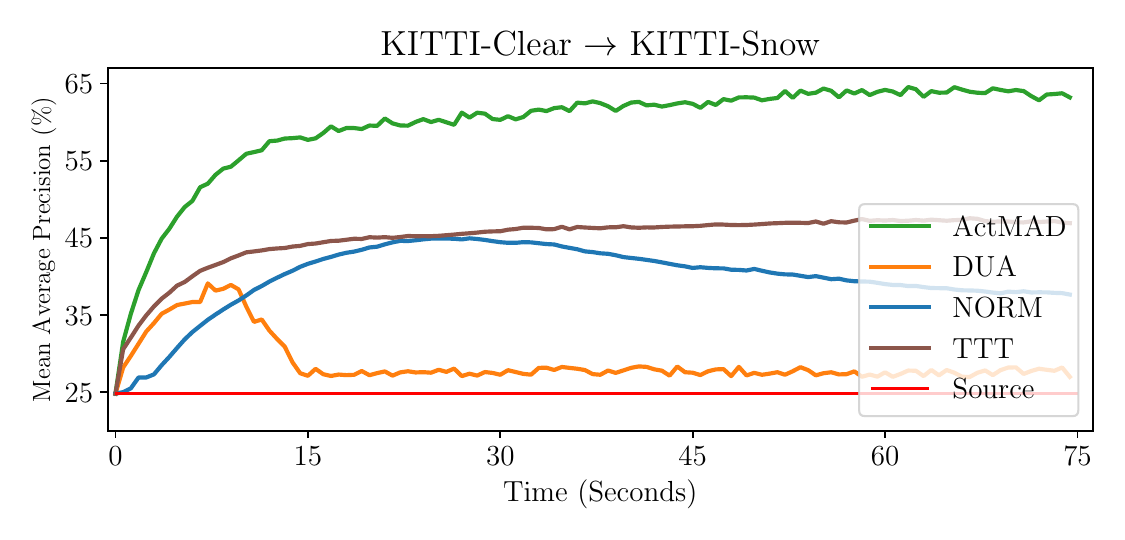}
        \vspace{-0.23cm}
        \caption{Online adaptation for each individual weather condition and comparison with baselines. 
        We again report the Mean Average Precision (mAP@50) averaged over all the 8 classes in the KITTI dataset.}
        \label{fig:online-adaptation-kitti}
\end{minipage}
\vspace{-0.1cm}
\end{figure*}

Results for CIFAR-10/100C for WRN-40-2 are reported in Table~\ref{tab:cifar-10C-results} and~\ref{tab:cifar-100C-results}, while the ViT results are listed in Table~\ref{tab:cifar-10-100C-results-vit}. 
ActMAD outperforms other online methods in classifying CIFAR-10C and CIFAR-100C images by a fair margin, establishing a new state-of-the-art for both Transformer-based and convolution\mk{al} architectures.
\mk{
When adapting the wide ResNet to CIFAR-10C, there is almost no gap between ActMAD and the offline approaches, while on CIFAR-100C offline adaptation fares better.
}
\mk{For the ViT, we also achieved a new state-of-the art on CIFAR-10C and CIFAR-100C, outperforming even CFA~\cite{kojima2022cfa}, designed specifically for robustifying vision transformers.}
\MK{I removed the numbers because they did not check with the Table NB, the unit of difference of percents is percent points (pp, or p.p.), not percents.
Or do you report relative error difference? I personally find this confusing.}

The results for the ImageNet-C dataset are shown in Table~\ref{tab:imagenet-c-results}.  ActMAD comes in second after CFA~\cite{kojima2022cfa}, but the performance gap in classification error is only $0.1$ percent-point. We suspect the advantage of CFA to stem from its alignment of class-conditioned statistics, which may become more important for the rich and diverse set of ImageNet classes than for the limited sets of classes of the CIFAR datasets. However, the use of class-conditional statistics limits CFA to \mk{the classification task}.
Additionally, ActMAD can also be combined with the entropy based methods to obtain further gains, these results and comparison with EATA~\cite{eata} are provided in the supplementary. 
\vspace{-0.3cm}
\paragraph{Continuous adaptation to pertubations.}
Wang~\etal evaluated their TTT method CoTTA~\cite{wang2022cotta} by continuously adapting to different corruptions, instead of adapting to only a single corruption at a time. 
We tested ActMAD and other baselines in such a scenario on CIFAR-10/100C.
The results, listed in Table~\ref{tab:cifar-10-100C-results-cont-adapt}, show that ActMAD performs on par with CoTTA, and outperforms other baselines. 

\subsection{Object Detection}
For object detection experiments, we adapted a KITTI-pretrained YOLOv3~\cite{redmon2018yolov3} to degrading weather conditions, such as fog, rain, and snow.
\mk{We used a learning rate of $1\mathrm{e}{-4}$, and a batch size of $30$.}
We compared our object detection results with DUA, NORM, and TTT. 
\begin{table*}
\small
\center
\setlength{\tabcolsep}{4pt}
\begin{tabular}{llcc|cc}
                                       &                              & \multicolumn{2}{c}{CIFAR-10C}              & \multicolumn{2}{c}{KITTI-FOG} \\
                                       &                              & \% Error $\downarrow$ & Change   & mAP@50 $\uparrow$ & Change  \\
\midrule
\multicolumn{2}{l}{Source (no adaptation) }                                     &   18.3 &         &   19.6 &  \\
\midrule
\multicolumn{2}{l}{Full ActMAD}                              &   10.4 &              0&   47.7 &0    \\
    \qquad Replace\quad\emph{multi-layer alignment}       & by\quad\emph{last layer alignment}       &   12.3 & +1.9         &   36.0 & -11.7   \\
    \qquad Replace\quad\emph{pixel statistics}            & by\quad\emph{channel averaged statistics}         &   11.5 & +1.1          &   38.6 & - 9.1   \\
    \qquad Replace\quad\emph{mean and variance alignment} & by\quad\emph{central moment difference}  &   10.5 & +0.1         &   41.2 & - 6.5   \\
    \qquad Replace\quad\emph{full parameter update}       & by\quad\emph{only affine parameter update}    &   10.6 & +0.2         &   45.2 & - 2.5   \\
\midrule
\end{tabular}
\caption{
\label{tab:ablation_alternatives}
Ablation study on design choices of ActMAD. We use a Wide-ResNet-40-2 for CIFAR-10C, and report the mean performance over 10 experiments for each of the 15 corruptions. We use YOLOv3 for the KITTI-FOG experiment. 
}
\end{table*}
\mk{
DUA and NORM could be applied to detector adaptation without any modifications,
thanks to the fact that they do not rely on the network to produce normalized probabilities and are not limited to adapting models trained in a multi-task setting.
By contrast, to deploy TTT, we had to re-train the detector with the auxiliary, self-supervised rotation prediction task.
We describe the implementation details in the supplemental.
}
The results, reported in Table~\ref{tab:kitti_fog_rain_snow}, show that ActMAD outperforms the baselines by a significant margin.
Additionally, in Figure~\ref{fig:online-adaptation-kitti} we plot the performance of the network during adaptation, showing that ActMAD stabilizes its precision as adaptation progresses, while the baselines provoke performance fluctuations. 
Needless to say, consistent level of performance is of utmost importance in safety-critical applications, like autonomous driving.
Further, for comparison with MemCLR~\cite{memclr} and results of ActMAD in other realistic scenarios, we refer to the supplementary material.

\subsection{Ablation Studies}
\label{sec:ablations}
\paragraph{Effect of batch size.}
\begin{table}
    \centering
    \small
\setlength{\tabcolsep}{3.0pt}
    \begin{tabular}{c|cccccc|c}
         BS& 125&100&75&50&25&10&\emph{Source Only}  \\
         \midrule
         Error $\downarrow$ (\%)&66.0&66.4&68.9&70.5&71.7&72.8&86.2\\ 
    \end{tabular}
    \caption{Top-1 Error (\%) averaged over $15$ corruptions in the ImageNet-C dataset while adapting an ImageNet pre-trained ResNet-18 with different batch sizes (BS).}
    \label{fig:batch_size_ablation}
\end{table}
The number of samples used by ActMAD to compute the location-aware statistics is equal to the number of batch elements, and therefore much lower than the number of samples used to compute the channel-wise statistics in competing methods.
Statistics computed on small samples have higher variance, which could adversely affect adaptation performance.
We investigate how decreasing the batch size affects the performance of ActMAD.
Like in the image classification experiments, we use an ImageNet pre-trained ResNet-18 from the PyTorch model zoo.
We decrease the learning rate in proportion to the decreasing batch size, following~Goyal~\etal~\cite{goyal2017accurate}.
As shown in Table~\ref{fig:batch_size_ablation},
\mk{limiting the batch size decreases the performance gently, and even batches as small as 10 images still let us achieve 13.4 percent-point improvement over the un-adapted model.}
\MK{\sout{We therefore conclude that it is safe to run ActMAD on batches as small as $10$. 
This corresponds to less than one second of a video. 
Given that ActMAD needs between a few and thirty seconds to attain high performance, as reported in Fig.~\ref{fig:online-adaptation-kitti}, the time needed to collect a batch of images does not constitute a significant delay.
}}
\MK{I removed this argument, as with the new numbers it is no longer true. ActMAD is competitive up to BS of 75, but this s already 3 seconds.}
\vspace{-0.25cm}
\paragraph{Design alternatives.}
\mk{To verify how our design choices contribute to the performance of ActMAD, we removed each of its key features and evaluated the resulting algorithms on image classification and object detection.
We adapted a wide ResNet from CIFAR-10 to CIFAR-10C, and YOLOv3 from KITTI to KITTI-Fog, in the same setup as in previous experiments.
}
As can be seen in Table~\ref{tab:ablation_alternatives}, 
\mk{a large} performance 
drop was \mk{provoked} by
limiting the alignment to the last feature map, as opposed to multiple feature maps throughout the network, and by aligning channel, instead of pixel, statistics.
This attests to the importance of fine-grained supervision, distributed over multiple layers and locations in the feature maps. 
Unsurprisingly, the use of location-aware statistics results in higher performance gain for KITTI~($\sim10$ percent-point) than for CIFAR~($\sim1$ percent-point). 
This is expected, since road scenes of KITTI are much more structured than the CIFAR images. 
\mk{We also hypothesize that ActMAD's localized feature alignment benefits object detection more, because detection relies on local features to a higher degree than the location-agnostic image classification.}
The type of statistics used for feature alignment seems to make no difference for CIFAR-10C, but aligning means and variances gives better results than aligning higher-order central moments for the KITTI object detector. This difference might stem from the fact that the KITTI experiments use almost ten times smaller batch size, and that low batch size affects higher-order moment estimates more than estimates of means and variances.
Finally, limiting the parameter update to channel-wise affine transformations results in a small performance drop in object detection, while almost not affecting the image classification accuracy. 
However, since we observed no drift in the continuous adaptation experiment and limiting the update might require manually inserting affine layers to the network, we prefer to perform the full update.

\section{Conclusion}
We propose a new method -- ActMAD -- for online test-time-training based on location-aware feature alignment. In contrast to many previous TTT approaches, our ActMAD \mk{is task- and architecture-agnostic. It} goes beyond image classification and can be readily applied to object detection. ActMAD outperforms existing approaches by a considerable margin, while being more stable and adapting faster \mk{to new distributions}. The power of ActMAD stems from fine-grained supervision which results from individually aligning each pixel in multiple feature maps across the network. \mk{Experiments} show that the fine-grained supervision provided by ActMAD from location-aware feature alignment is especially helpful for tasks which require a holistic understanding of the scene, for example object detection. 
\section*{Acknowledgment}
We gratefully acknowledge the financial support by the Austrian Federal Ministry for Digital and Economic Affairs, the National Foundation for Research, Technology and Development and the Christian Doppler Research Association. 
This work was also partially funded by the FWF Austrain Science Fund Lise Meitner grant (M3374) and Austrian Research Promotion Agency (FFG) under the projects High-Scene (884306) and SAFER (894164).

\appendix
\section*{Supplementary Material}
\begin{figure*}
\begin{subfigure}{0.5\textwidth}
    \centering
    \includegraphics[scale=0.5,trim = 0 0 0 0, clip]{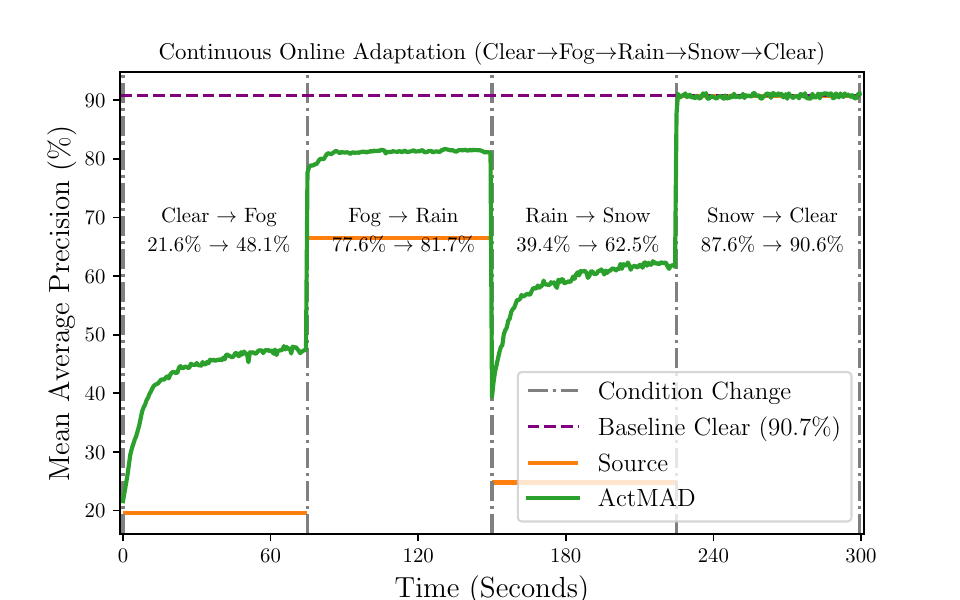}
    \caption{Clear$\to$Fog$\to$Snow$\to$Rain$\to$Clear}
    \label{fig:supp-day-fog-rain-snow-day}
    \end{subfigure}
\begin{subfigure}{0.5\textwidth}
    \centering
    \includegraphics[scale=0.5,trim = 0 0 0 0, clip]{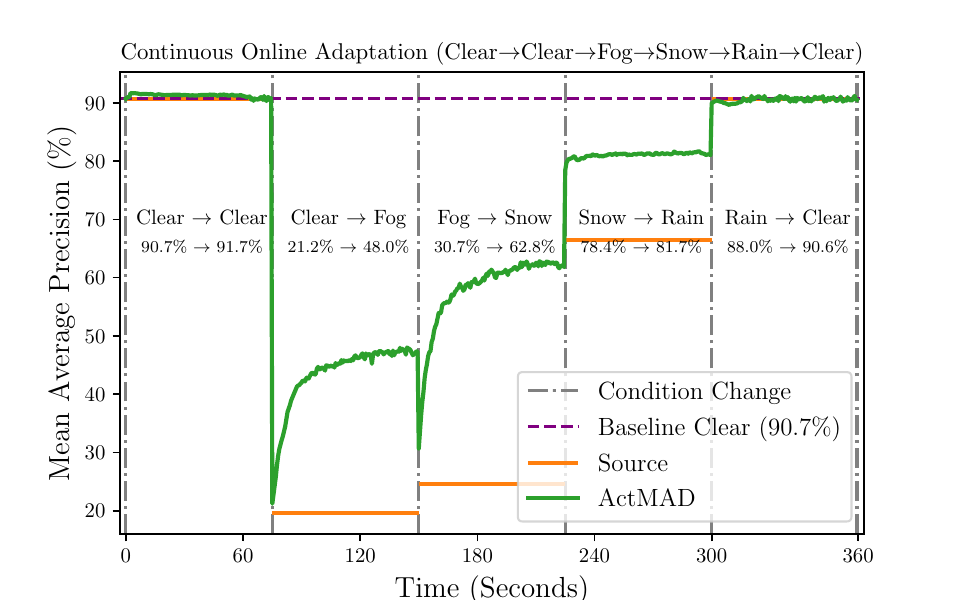}
    \caption{Clear$\to$Clear$\to$Fog$\to$Snow$\to$Rain$\to$Clear}
    \label{fig:supp-day-day-fog-rain-snow-day}
    \end{subfigure}
    \caption{Mean Average Precision (mAP@50) for two simulated continuous online adaptation scenarios with different weather sequences. \emph{Source} refers to the model trained on clear weather condition and tested on the changing weather without adaptation. }
\end{figure*}

\begin{figure*}
\begin{minipage}{0.48\textwidth}
    \centering
    \includegraphics[scale=0.45,trim = 10 15 10 10, clip]{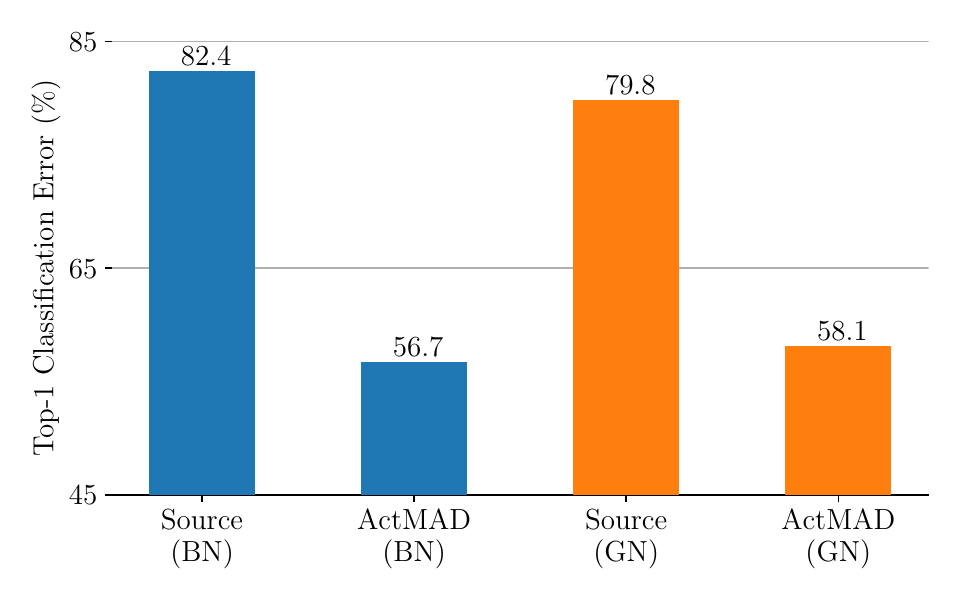}
    \caption{Mean Top-1 Classification Error (\%) over all corruptions in ImageNet-C (Level 5) using a ResNet-50 with different normalization layers. BN: Batch Normalization, GN: Group~Normalization. \emph{Source} refer to results obtained without adaptation.}
    \label{fig:supp-gn-bn}
    \end{minipage}
    \hfill
\begin{minipage}{0.48\textwidth}
    \centering
    \includegraphics[scale=0.45,trim = 10 15 10 10, clip]{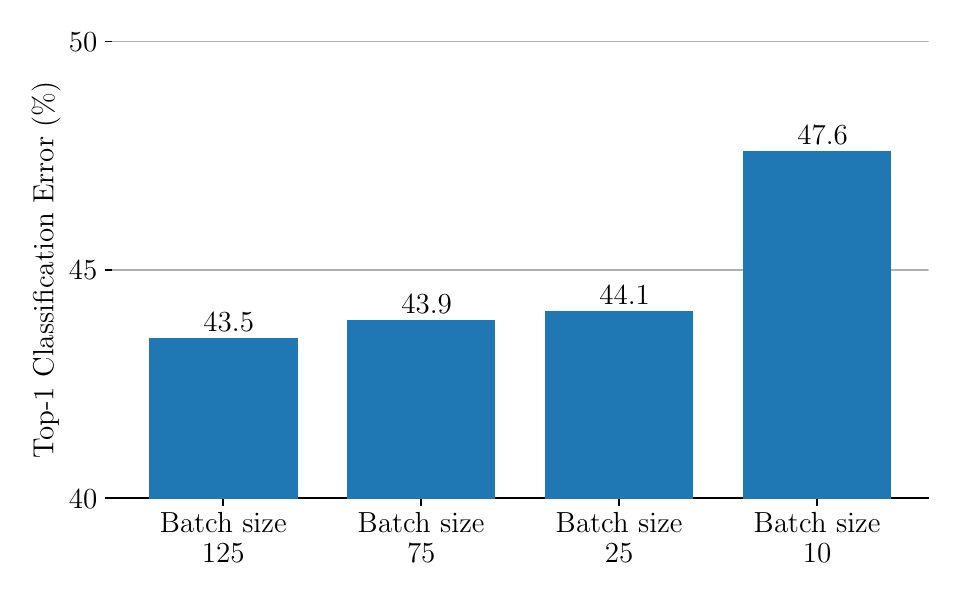}
    \caption{Mean Top-1 Classification Error (\%) over all corruptions in ImageNet-C (Level 5) while decreasing the batch size. The backbone used is an AugMix~\cite{hendrycks2019augmix} pre-trained ResNet-50. Here, \emph{Source} (without adaptation) Error is 61.1\%.}
    \label{fig:supp-bs-deepaug}
    \end{minipage}
\end{figure*}

\begin{table*}
\setlength\tabcolsep{4.0pt}
\centering
\small
\begin{tabular}{@{}l@{\hspace{2em}}ccccccccccccccc|l@{\hspace{0.1\tabcolsep}}l}
     \multicolumn{1}{r}{Corruptions:}            &  Gauss &  Shot  &  Impul &  Defcs &   Gls  &   Mtn  &   Zm   &   Snw  &  Frst  &   Fg   &  Brt   &   Cnt  &   Els  &   Px   &   Jpg  &  Mean & \\
      \cmidrule{0-17}
      \multicolumn{17}{c}{Level 4}\\
      \midrule
      Source & {24.1} & {17.1} & {16.4} & {6.6} & {23.5} & {8.4} & {7.4} & {12.2} & {11.5} & {8.3} & {6.2} & {9.2} & {10.6} & {19.4} & {13.1} & {12.9} \\
ActMAD & {11.7} &  {9.8} & {11.7} & {5.9} & {16.0} & {7.1} & {6.2} &  {9.1} &  {7.9} & {6.5} & {5.4} & {6.1} &  {9.2} &  {7.1} & {11.8} &  {8.8} &$\pm\,7\mathrm{e}-3$\\
      \cmidrule{0-17}
      \multicolumn{17}{c}{Level 3}\\
      \midrule
      Source & {20.4} & {14.6} & {9.7} & {5.4} & {12.9} & {8.6} & {6.5} & {9.9} & {11.4} & {6.3} & {5.5} & {7.2} & {7.4} & {9.6} & {12.1} & {9.8} \\
ActMAD & {10.5} &  {8.8} & {8.7} & {5.3} & {10.6} & {7.2} & {5.9} & {8.0} &  {7.9} & {5.9} & {5.1} & {5.6} & {6.9} & {6.1} & {10.9} & {7.6}&$\pm\,7\mathrm{e}-3$ \\
      \cmidrule{0-17}
      \multicolumn{17}{c}{Level 2}\\
      \midrule
      Source & {13.4} & {8.8} & {8.0} & {5.1} & {14.2} & {6.5} & {5.8} & {9.2} & {8.5} & {5.3} & {5.3} & {6.1} & {6.5} & {7.8} & {10.9} & {8.1} \\
ActMAD &  {8.4} & {6.6} & {7.3} & {5.0} & {10.9} & {6.2} & {5.6} & {7.1} & {6.7} & {5.1} & {5.0} & {5.3} & {6.4} & {5.8} & {10.0} & {6.8} &$\pm\,8\mathrm{e}-3$\\
      \cmidrule{0-17}
      \multicolumn{17}{c}{Level 1}\\
      \midrule
      Source & {8.7} & {6.5} & {6.2} & {4.9} & {14.1} & {5.5} & {5.9} & {6.4} & {6.5} & {4.9} & {5.0} & {5.0} & {6.9} & {5.8} & {8.7} & {6.7} \\
ActMAD & {6.4} & {5.7} & {6.0} & {4.9} & {10.3} & {5.5} & {5.6} & {5.9} & {5.6} & {4.8} & {4.9} & {4.9} & {6.8} & {5.3} & {7.8} & {6.0}&$\pm\,9\mathrm{e}-3$ \\
      \bottomrule
\end{tabular}
\caption{Top-1 Classification Error (\%) for all corruptions in CIFAR-10C (level 4 -- 1), highest severity is reported in the main manuscript. Lower is better. The results were obtained by adapting a \textbf{WRN-40-2 backbone}, trained on CIFAR-10, to CIFAR-10C. ActMAD results are averaged over 10 runs, and we report the mean error and its standard deviation.}

\label{tab:supp-cifar-10C-results}

\end{table*}

\begin{table*}
\setlength\tabcolsep{4.0pt}
\centering
\small
\begin{tabular}{@{}l@{\hspace{2em}}ccccccccccccccc|l@{\hspace{0.1\tabcolsep}}l}
     \multicolumn{1}{r}{Corruptions:}            &  Gauss &  Shot  &  Impul &  Defcs &   Gls  &   Mtn  &   Zm   &   Snw  &  Frst  &   Fg   &  Brt   &   Cnt  &   Els  &   Px   &   Jpg  &  Mean & \\
      \cmidrule{0-17}
      \multicolumn{17}{c}{Level 4}\\
      \midrule
      Source & {60.7} & {51.6} & {47.9} & {27.1} & {54.4} & {30.3} & {28.9} & {37.4} & {39.0} & {35.4} & {27.2} & {35.9} & {34.4} & {39.0} & {40.1} & {39.3} \\
ActMAD & {38.0} & {34.9} & {35.3} & {26.1} & {41.1} & {28.0} & {27.1} & {32.8} & {30.8} & {31.0} & {24.7} & {26.6} & {31.9} & {28.4} & {37.2} & {31.6} &$\pm\,4\mathrm{e}-3$\\
      \cmidrule{0-17}
      \multicolumn{17}{c}{Level 3}\\
      \midrule
      Source & {55.2} & {45.9} & {36.9} & {25.7} & {39.9} & {30.5} & {27.4} & {33.3} & {38.1} & {29.5} & {25.5} & {30.5} & {28.6} & {30.3} & {38.0} & {34.4} \\
ActMAD & {36.6} & {32.9} & {30.5} & {24.8} & {34.1} & {28.2} & {26.3} & {30.3} & {30.4} & {27.9} & {24.1} & {25.5} & {27.6} & {26.4} & {35.7} & {29.4}&$\pm\,3\mathrm{e}-3$ \\
      \cmidrule{0-17}
      \multicolumn{17}{c}{Level 2}\\
      \midrule
      Source & {44.6} & {34.5} & {30.7} & {24.3} & {41.5} & {27.7} & {26.2} & {32.7} & {31.8} & {26.8} & {24.4} & {27.5} & {27.9} & {28.0} & {36.5} & {31.0} \\
ActMAD & {32.4} & {28.5} & {27.4} & {24.0} & {33.7} & {26.5} & {25.4} & {28.9} & {28.0} & {25.4} & {23.7} & {24.8} & {27.6} & {26.1} & {34.4} & {27.8}&$\pm\,2\mathrm{e}-3$ \\
      \cmidrule{0-17}
      \multicolumn{17}{c}{Level 1}\\
      \midrule
      Source & {34.4} & {29.6} & {26.9} & {23.8} & {42.9} & {25.6} & {26.1} & {26.1} & {27.4} & {24.0} & {23.8} & {24.3} & {28.4} & {25.2} & {32.4} & {28.1} \\
ActMAD & {28.5} & {26.7} & {25.6} & {23.5} & {33.7} & {24.9} & {25.1} & {25.4} & {25.6} & {23.7} & {23.5} & {23.8} & {28.2} & {24.8} & {31.0} & {26.3}&$\pm\,2\mathrm{e}-3$ \\
      \bottomrule
\end{tabular}
\caption{Top-1 Classification Error (\%) for all corruptions in CIFAR-100C (level 4 -- 1), highest severity is reported in the main manuscript. Lower is better. The results were obtained by adapting a \textbf{WRN-40-2 backbone}, trained on CIFAR-100, to CIFAR-100C. ActMAD results are averaged over 10 runs, and we report the mean error and its standard deviation.}

\label{tab:supp-cifar-100C-results}

\end{table*}

\begin{table*}
\setlength\tabcolsep{4.0pt}
\centering
\small
\begin{tabular}{@{}l@{\hspace{2em}}ccccccccccccccc|l@{\hspace{0.1\tabcolsep}}l}
     \multicolumn{1}{r}{Corruptions:}&  Gauss &  Shot  &  Impul &  Defcs &   Gls  &   Mtn  &   Zm   &   Snw  &  Frst  &   Fg   &  Brt   &   Cnt  &   Els  &   Px   &   Jpg  &  Mean & \\
      \cmidrule{0-17}
      \multicolumn{17}{c}{Level 4}\\
      \midrule
      Source & {93.2} & {94.7} & {94.3} & {84.5} & {89.4} & {85.3} & {77.2} & {83.4} & {79.4} & {72.8} & {44.5} & {88.1} & {63.4} & {71.2} & {58.8} & {78.7} \\
ActMAD & {65.7} & {70.1} & {68.6} & {68.4} & {67.0} & {64.8} & {58.4} & {64.6} & {60.7} & {48.3} & {37.7} & {57.3} & {41.9} & {48.1} & {48.4} & {58.0} $\pm\,2\mathrm{e}-2$\\
      \cmidrule{0-17}
      \multicolumn{17}{c}{Level 3}\\
      \midrule
      Source & {80.9} & {82.7} & {82.9} & {74.1} & {85.4} & {73.9} & {71.9} & {73.6} & {77.8} & {66.2} & {39.6} & {65.3} & {51.1} & {56.7} & {49.3} & {68.8} \\
ActMAD & {55.2} & {56.4} & {57.7} & {58.8} & {61.5} & {53.3} & {54.0} & {56.6} & {59.7} & {45.3} & {35.2} & {43.5} & {37.9} & {41.0} & {41.5} & {50.5} $\pm\,1\mathrm{e}-2$\\
      \cmidrule{0-17}
      \multicolumn{17}{c}{Level 2}\\
      \midrule
      Source & {63.6} & {67.8} & {74.6} & {59.4} & {65.4} & {57.7} & {65.2} & {77.5} & {67.0} & {56.3} & {36.4} & {51.5} & {60.9} & {43.3} & {46.3} & {59.5} \\
ActMAD & {46.7} & {48.3} & {52.8} & {47.9} & {47.5} & {43.7} & {49.7} & {56.5} & {52.0} & {41.4} & {33.4} & {38.6} & {52.0} & {35.8} & {39.3} & {45.7}$\pm\,9\mathrm{e}-3$ \\
      \cmidrule{0-17}
      \multicolumn{17}{c}{Level 1}\\
      \midrule
      Source & {50.5} & {53.1} & {61.9} & {51.3} & {52.4} & {45.2} & {55.9} & {55.5} & {49.7} & {49.0} & {34.2} & {44.0} & {40.4} & {41.3} & {42.6} & {48.5} \\
ActMAD & {40.6} & {41.4} & {46.6} & {42.2} & {40.6} & {37.7} & {44.6} & {44.1} & {41.4} & {38.8} & {32.1} & {36.2} & {37.8} & {35.1} & {36.9} & {39.7}$\pm\,6\mathrm{e}-3$ \\
      \bottomrule
\end{tabular}
\caption{Top-1 Classification Error (\%) for all corruptions in ImageNet-C (level 4 -- 1), highest severity is reported in the main manuscript. Lower is better. The results were obtained by adapting a \textbf{ResNet-18 backbone}, trained on ImageNet, to ImageNet-C. ActMAD results are averaged over 10 runs, and we report the mean error and its standard deviation.}

\label{tab:supp-imagenet-C-results}

\end{table*}

\begin{table*}
\setlength\tabcolsep{4.0pt}
\centering
\small
\begin{tabular}{@{}l@{\hspace{2em}}ccccccccccccccc|l@{\hspace{0.1\tabcolsep}}l}
      \multicolumn{1}{r}{Corruptions:}            &  Gauss &  Shot  &  Impul &  Defcs &   Gls  &   Mtn  &   Zm   &   Snw  &  Frst  &   Fg   &  Brt   &   Cnt  &   Els  &   Px   &   Jpg  &  Mean \\
      \cmidrule{0-17}
      \multicolumn{17}{c}{CIFAR-10C}\\
      \midrule
      Source & {23.6} & {21.9} & {28.7} & {5.2} & {27.9} & {9.1} & {4.4} & {5.5} & {8.4} & {14.2} & {2.4} & {15.2} & {19.9} & {25.5} & {13.3} & {15.0} \\
ActMAD & 12.9 &11.0& 11.7&  4.1& 11.2&  5.8& 2.7&  4.5&  4.4&
        9.3&  2.0&  6.1&  7.2&  4.4& 10.9&7.3&$\pm\,0.1$\\
      \cmidrule{0-17}
      \multicolumn{17}{c}{CIFAR-100C}\\
      \midrule
      Source & {55.0} & {52.9} & {57.8} & {18.0} & {60.5} & {23.6} & {16.0} & {22.3} & {27.5} & {34.2} & {11.9} & {35.3} & {34.8} & {43.3} & {33.7} & {35.1} \\
ActMAD  &31.8&29.7&28.5&18.5&28.3&20.1&14.3&19.2&17.9&26.8&11.9&21.3&23.7&15.4&29.5&22.4&$\pm\,0.1$\\
\midrule
            \multicolumn{17}{c}{ImageNet-C}\\
      \midrule
Source&53.1& 52.4& 53.1& 57.3&65.8&49.6&55.3&43.1&47.4& 43.5& 23.9&
       68.2& 53.3& 34.5& 34.0& 49.0\\
       \midrule
       ActMAD&44.0& 41.9& 42.4& 47.7& 46.9& 40.5& 40.3& 32.6& 36.2& 31.5& 20.6&
       39.0& 33.4& 27.0 &30.6&36.9&$\pm\,0.3$\\
      \bottomrule
\end{tabular}
\caption{Top-1 Classification Error (\%) for all corruptions in CIFAR-10/100C and ImageNet-C (level 5). Lower is better. All results are obtained by using a \textbf{ViT-B/16 backbone}.  ActMAD results are averaged over 10 runs, and we report the mean error and its standard deviation.}
\label{tab:supp-cifar-10-100C-results-ViT}
\end{table*}

\begin{table*}
\setlength\tabcolsep{4.0pt}
\centering
\small
\begin{tabular}{@{}l@{\hspace{2em}}ccccccccccccccc|l@{\hspace{0.1\tabcolsep}}l}
      \multicolumn{1}{r}{Corruptions:}            &  Gauss &  Shot  &  Impul &  Defcs &   Gls  &   Mtn  &   Zm   &   Snw  &  Frst  &   Fg   &  Brt   &   Cnt  &   Els  &   Px   &   Jpg  &  Mean \\
      \cmidrule{0-17}
      \multicolumn{17}{c}{CIFAR-10C}\\
      \midrule
      Source      & {28.8} & {22.9} & {26.2} & { 9.5} & {20.6} & {10.6} & { 9.3} & {14.2} & {15.3} & {17.5} & { 7.6} & {20.9} & {14.7} & {41.3} & {14.7} & {18.3} & \\
CoTTA &  {\bfseries{12.9}} & {\underline{13.8}} &  {\bfseries{12.0}} & {\underline{9.1}} &  {\bfseries{14.3}} &  {\bfseries{9.2}} & {\underline{9.2}} & {\underline{12.0}} & {\bfseries{10.0}} &  {\bfseries{12.1}} & {\underline{7.0}} & {\underline{14.3}} & {\underline{14.1}} & {\bfseries{9.9}} &  {\bfseries{12.7}} & {\underline{11.5}} &$\pm\,0.1$\\
ActMAD & {\underline{13.3}} &  {\bfseries{11.6}} & {\underline{15.7}} &  {\bfseries{8.7}} & {\underline{16.7}} & {\underline{9.9}} &  {\bfseries{8.3}} &  {\bfseries{10.5}} & {\bfseries{10.0}} & {\underline{12.8}} &  {\bfseries{6.9}} &  {\bfseries{10.6}} &  {\bfseries{13.3}} & {\bfseries{9.9}} & {\underline{13.4}} &  {\bfseries{11.4}} &$\pm\,0.1$ \\

      \cmidrule{0-17}
      \multicolumn{17}{c}{CIFAR-100C}\\
      \midrule
Source &             {65.7} &             {60.1} &             {59.1} &             {32.0} &             {51.0} &             {33.6} &             {32.4} &             {41.4} &             {45.2} &             {51.4} &             {31.6} &             {55.5} &             {40.3} &             {59.7} &             {42.4} &             {46.8}\\
 CoTTA & {\underline{40.8}} & {\underline{39.2}} &  {\bfseries{40.8}} &  {\bfseries{29.9}} &  {\bfseries{40.9}} &  {\bfseries{30.7}} &  {\bfseries{29.8}} & {\underline{35.9}} & {\underline{34.9}} & {\underline{43.4}} &  {\bfseries{26.9}} & {\underline{38.2}} &  {\bfseries{36.6}} &  {\bfseries{30.9}} &  {\bfseries{36.8}} &  {\bfseries{35.7}}&$\pm\,0.4$  \\
ActMAD &  {\bfseries{39.6}} &  {\bfseries{38.3}} & {\underline{41.4}} & {\underline{31.8}} & {\underline{41.7}} & {\underline{32.7}} & {\underline{30.9}} &  {\bfseries{34.6}} &  {\bfseries{33.6}} &  {\bfseries{42.9}} & {\underline{27.4}} &  {\bfseries{36.3}} & {\underline{37.2}} & {\underline{31.9}} & {\underline{38.6}} & {\underline{35.9}} &$\pm\,0.5$\\

\bottomrule

\end{tabular}
\caption{Top-1 Classification Error (\%) for all corruptions in CIFAR-10/100C (level 5), while continuously adapting to corruptions. 
Lower is better. 
These results were obtained by adapting a \textbf{Wide-ResNet-40-2 backbone}, trained on CIFAR-10/100, to CIFAR-10/100C. ActMAD results are averaged over 10 runs, and we report the mean error and its standard deviation. The lowest error is shown in bold, the second best is underlined.}
\label{tab:supp-randomizing-corruptions}
\end{table*}

\begin{table*}
\begin{minipage}{0.47\textwidth}
\small
\setlength{\tabcolsep}{3.0pt}
\centering
 \begin{tabular}{c c c | c c}
          TENT & EATA  & ActMAD & TENT+ActMAD & EATA+ActMAD       \\
          \midrule
    67.2 & {\bf64.9}  &   66.0 &     64.1    & {\bf62.7}    \\
    \bottomrule
    \end{tabular}
    \caption{Top-1 Error (\%) averaged over $15$ corruptions in the ImageNet-C dataset while adapting an ImageNet pre-trained ResNet-18 from PyTorch model zoo.}
    \label{tab:actmad-entropy}
\end{minipage}\hfill
\begin{minipage}{0.47\textwidth}
\setlength{\tabcolsep}{2.0pt}
\small
\centering
\begin{tabular}{l c c c}
        & \emph{Source-only} & MemCLR & ActMAD \\
        \midrule
    mean Avg. Prec. (mAP) & 25.2  &  29.8  &   {\bf34.5} \\	
    \bottomrule
    \end{tabular}
    \caption{Mean Average Precision (mAP@50) for a Faster R-CNN pre-trained on the Cityscapes dataset and adapted to the FoggyCityScapes test split.}
    \label{tab:actmad-memclr}
\end{minipage}
\label{tab:ablation-datasetsize-batchsize}
\end{table*}

\begin{table}[]
    \centering
    \small
    \setlength{\tabcolsep}{2.0pt}
    \begin{tabular}{l c c c c c c}
        CFA &  DUA & TENT & SHOT& TTT++ & ActMAD$^{\text{full}}$ & ActMAD$^{\text{affine}}$ \\
        \midrule
        5.3 &  2.1 & 5.2  & 6.1 & 6.8   & 5.8    & 5.2 \\
        \bottomrule
    \end{tabular}
    \caption{Total runtime (seconds) for CIFAR-10C, batch size 128 (see Table~1, main paper). ActMAD$^{\text{affine}}$ results are provided in Table~7, main paper.}
    \label{tab:run-time}
\end{table}
In the following, we present additional continuous online adaptation scenarios in which we test ActMAD (Section~\ref{sec:cont-adapt}), provide all implementation details for baselines and our experiments to support reproducibility (Section~\ref{sec:implementation-details}), present additional ablation studies (Section~\ref{sec:additional-results}), and more comparisons and insights for our ActMAD (Section~\ref{sec:additional-insights}).    
\section{Continuous Online Adaptation}
\label{sec:cont-adapt}
Our location-aware feature alignment 
is especially helpful for the task of object detection. 
The fine-grained supervision helps ActMAD to adapt to different weather scenarios in a dynamic and efficient manner. 
Moreover, ActMAD also shows promising results while adapting to different weather scenarios in a continuous manner.
In the main manuscript (Figure 1), we show such a simulated continuous adaptation scenario (Clear$\to$Fog$\to$Snow$\to$Rain$\to$Clear). 
We further test ActMAD in similar scenarios where the weather conditions are encountered in a different sequence. 
In Figure~\ref{fig:supp-day-fog-rain-snow-day} and~\ref{fig:supp-day-day-fog-rain-snow-day}, we present results for adaptation with ActMAD in two such scenarios. 

The results show that ActMAD can achieve impressive results in different scenarios and the sequence of weather conditions which are encountered does not matter. 
This is close to the requirements in the real-world, where different weather conditions can occur in any order during driving. 
We also see from the results that ActMAD can even improve the performance of a KITTI~\cite{geiger2013vision} pre-trained YOLOv3~\cite{redmon2018yolov3} on the validation split from the dataset (which represents the \textit{Clear} weather condition). 
For example, in Figure~\ref{fig:supp-day-day-fog-rain-snow-day}, where order of the weather conditions is \textit{Clear$\to$Clear$\to$Fog$\to$Snow$\to$Rain$\to$Clear}, ActMAD achieves $1.0\%$ higher Mean Average Precision while adapting from \textit{Clear$\to$Clear} at the beginning of the adaptation cycle.
This is because even though the adaptation is performed from~\textit{Clear$\to$Clear}, the activation statistics could be different between the training and inference,~\eg, due to slight change in the environmental conditions or scenery.  
Our ActMAD helps to rectify this mismatch to obtain an increase in performance. 

\section{Implementation Details}
\label{sec:implementation-details}
\paragraph{Hardware:} All our experiments are performed on a single NVIDIA\textsuperscript{\textregistered} GeForce\textsuperscript{\textregistered} RTX 3090 
and a GPU server consisting of four Quadro RTX 8000.

\paragraph{CIFAR-10/100C:} We use an AugMix pre-trained~\cite{hendrycks2019augmix} Wide-Resnet-40~\cite{zagoruyko2016wide} for our main results on CIFAR-10/100C. 
For experiments with ViT-B/16~\cite{dosovitskiy2020image} we finetune the ViT backbone (pre-trained on ImageNet-21k) for 10000 iterations on the training sets of CIFAR-10/100~\cite{long2015learning} 
with a learning rate of ${3}\mathrm{e}{-2}$. 

\paragraph{ImageNet-C:} All ImageNet-C experiments use ImageNet pre-trained models. 
For the main results with ResNet-18~\cite{he2016deep} we use the pre-trained weights from the PyTorch~\cite{NEURIPS2019_9015} model zoo.
For experiments with ResNet-50 using Group Normalization~\cite{wu2018group} (Section~\ref{subsec:diff-norm-tech}), we use pre-trained weights from a third-party open source repository\footnote{\href{https://github.com/ppwwyyxx/GroupNorm-reproduce/tree/883f694}{github.com/ppwwyyxx/GroupNorm-reproduce},~commit: 883f694}.
For experiments with a DeepAug~\cite{hendrycks2021many} pre-trained ResNet-50 (Section~\ref{subsec:imagenet-c-augmix}), we use the pre-trained weights from Robust Bench\footnote{\label{robbench}\href{https://robustbench.github.io/}{robustbench.github.io/}}.
For experiments on ImageNet-C with ViT-B/16 model (Section~\ref{subsec:vit-results}), we use the pre-trained weights from the open source \emph{timm} library\footnote{\href{https://github.com/rwightman/pytorch-image-models}{github.com/rwightman/pytorch-image-models},~release: 0.6.11}. 

\paragraph{KITTI:} All KITTI~\cite{geiger2013vision} experiments start with a model pre-trained on the original KITTI dataset. 
As it is collected mostly in clear (\ie sunny) weather conditions, we refer to it as KITTI-Clear. 
We adapt this model to degrading weather conditions,~\ie fog~\cite{halder2019physics}, rain~\cite{halder2019physics} and snow~\cite{hendrycks2019robustness}. 
For KITTI-Clear, we re-train an MS-COCO~\cite{lin2014microsoft} pre-trained YOLOv3 in a supervised manner on the train split of the KITTI dataset. 
The model is trained for 100 epochs with a batch size of 30, while all the other training details are kept constant as in the PyTorch implementation of YOLOv3\footnote{\href{https://github.com/ultralytics/yolov3/tree/d353371}{github.com/ultralytics/yolov3}, commit: d353371}. 
\paragraph{Baselines:} Since all the baselines used to compare our results are primarily tested for object recognition on CIFAR-10/100C~\cite{hendrycks2019robustness} and ImageNet-C~\cite{hendrycks2019robustness}, we use their official open-source codes and use all the hyper-parameters, which they report in their paper.

For object detection experiments, only DUA~\cite{mirza2021dua} provides official results\footnote{\href{https://github.com/jmiemirza/DUA/tree/d6e5398}{https://github.com/jmiemirza/DUA}, commit: d6e5398}. 
Therefore, we implement NORM~\cite{schneider2020norm} for object detection using their official repository\footnote{\href{https://github.com/bethgelab/robustness/tree/aa0a679}{https://github.com/bethgelab/robustness},~commit: aa0a679}.
Since NORM is a gradient-free approach, it does not rely on any hyper-parameters. 
We use a batch size of 30 to obtain all the results for NORM.

TTT~\cite{sun2020ttt} uses rotation prediction as an auxiliary task for test-time training. 
TTT is also primarily tested for object recognition by its authors. 
We implement TTT for object detection for the purpose of comparison in our paper. 
We use the same batch size (\ie~30), as for ActMAD.
For joint-training with the auxiliary rotation prediction task we take the output from the feature encoder of YOLOv3 and use a single-layer linear classifier head to predict the rotation of the input image. 
The hyperparameters used for the KITTI pre-training (for ActMAD experiments) and joint-training are kept the same. 
However, we find that training a network jointly for the classification task of rotation prediction and object detection requires special design choices. 
For example, we use a loss scaling factor of $0.4$ for the self-supervised task during joint-training to make the network converge. 
Similarly, the same loss scaling factor is used during test-time training. 
The evaluation protocol is kept consistent for all baselines and our method.

We also implemented TTT++\cite{liu2021tttpp} on YOLOv3 in order to use it as a baseline for object detection experiments.
We used the same prediction head, hyper-parameters and the augmentations (for creating two augmented views for SimCLR~\cite{chen2020simple} training) used by the original TTT++ paper for their experiments on CIFAR-10/100. 
We experimented with several design choices for joint-training (object detection and contrastive SimCLR task), each for a total of $500$ epochs on the KITTI train set, starting from the pre-trained weights on MS-COCO. 
However, despite the best of our efforts we could not make the joint-training converge. 
A potential issue could be the requirement of using larger batch sizes for training the contrastive learning task. 
Initially, we tested with a batch size of $100$ for joint training by keeping the original aspect ratio of KITTI images~(\ie $370\,\times\,1224$). 
However, realizing that we might require larger batch sizes, 
we had to downscale the KITTI images equal to the aspect ratio of ImageNet images~(\ie $224\,\times\,224$), to be able to increase the batch size to $300$, but in this case the regression loss for object detection did not converge and we get low performance on the validation set of KITTI. 
We suspect that it might be because of the drastic decrease in the image size, as compared to the original aspect ratio of images in the KITTI dataset, which is unsuitable for the object detection task on the KITTI dataset. 


\section{Additional Results}
\label{sec:additional-results}
In this section we provide additional ablation studies, results on lower severity levels for ActMAD, detailed results for the vision transformer backbone~\cite{dosovitskiy2020image} and the Continuous Adaptation experiment.
\subsection{Different Normalization Techniques}
\label{subsec:diff-norm-tech}
Recent test-time training approaches which adapt the network statistics,~\eg NORM~\cite{wang2020tent} and DUA~\cite{mirza2021dua}, critically rely on batch normalization~\cite{ioffe2015batch}.
However, if the network is equipped with other forms of normalization,~\eg~Instance~\cite{ulyanov2016instance}, Layer~\cite{ba2016layer} or Group Normalization~\cite{wu2018group}, these approaches are not applicable.
In contrast, our ActMAD is agnostic to the architecture and can work with any type of normalization applied in the network. 
We already showed in the main manuscript (Table 3) that it can be applied to a Transformer-based architecture, which uses Layer Norm.
In order to test the performance of ActMAD with different normalization layers inside the convolution-based architectures, we test ActMAD with a ResNet-50 equipped with Group Normalization. 
Figure~\ref{fig:supp-gn-bn} shows the adaptation results of a ResNet-50 with both Batch Normalization~(BN) and Group Normalization~(GN).
Using GN, our ActMAD performance decreases by only $1.8\%$ over the same architecture using~BN.

\subsection{Robustifying an Already Robust Model}
\label{subsec:imagenet-c-augmix}
ActMAD can be effectively used with any off-the-shelf pre-trained model. 
To test this, we apply ActMAD for test-time adaptation of an already robust, AugMix~\cite{hendrycks2019augmix} pre-trained ResNet-50. 
We further decrease the batch size for adaptation on this backbone. 
Results for this experiment (highest severity) are provided in Figure~\ref{fig:supp-bs-deepaug}.
We find that ActMAD decreases the error significantly even for an already robust model.
For example, the Source (without adaptation) top-1 error (\%) over the $15$ corruptions on ImageNet-C is $61.1\%$, while after ActMAD adaptation the error decreases to merely $43.5\%$. 
Furthermore, we also see that while applying ActMAD to a robust model, the decrease in performance with decreasing batch size is also minimal (similar to the ablation study in the main manuscript, Figure 4, with ResNet-18 backbone). 
For example, while decreasing the batch size from $125$ to $10$, the error increases by only $4.1\%$. 
\subsection{Lower Severity Results}
\label{sec:supp-lower-severity-levels}
In the main manuscript we provide results for the highest severity (Level 5) in the Robust Bench~\cite{hendrycks2019robustness}. 
Our ActMAD achieves impressive gains in performance even for lower severity levels. 
In Table~\ref{tab:supp-cifar-10C-results} and Table~\ref{tab:supp-cifar-100C-results} we provide the mean and standard deviation of results (over $10$ random runs) obtained by ActMAD on lower severities~(\ie Levels 4 -- 1) on~CIFAR-10/100C. 
We use the Wide-ResNet-40 model for these evaluations. 
Furthermore, in Table~\ref{tab:supp-imagenet-C-results}, we provide the mean and standard deviation of results over $10$ random runs on the large scale ImageNet-C (with ResNet-18) dataset for lower severities (Levels 4 -- 1).
The results show that ActMAD can even be applied when the distribution shift between the train and test data is not too large. 
\begin{figure}
    \centering
    \includegraphics[width=.95\linewidth]{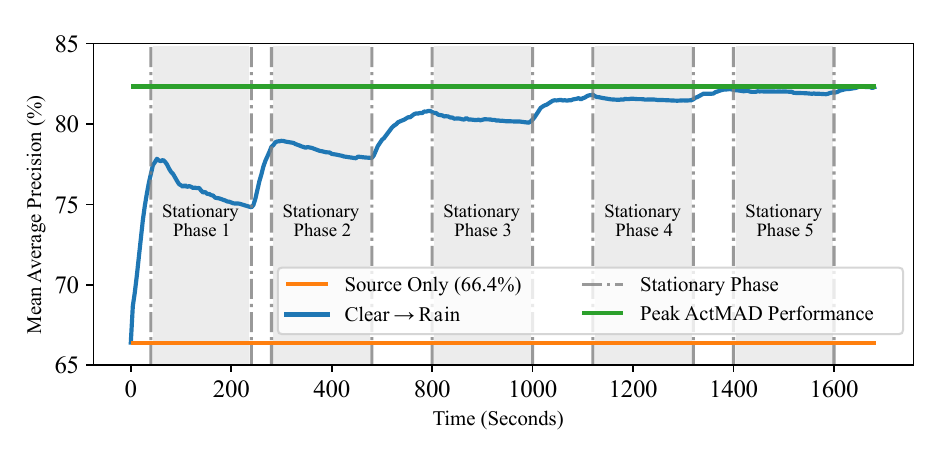}
    \caption{Mean Average Precision (mAP@50) for ActMAD adaptation from KITTI-Clear $\to$ KITTI-Rain for a simulated scenario where the car is assumed to be stationary. The gray patches represent the randomly chosen `stationary' intervals.}
    \label{fig:parking-mode}
\end{figure}
\subsection{Vision Transformer}
\label{subsec:vit-results}
ActMAD can also be seamlessly applied to the Vision Transformer backbone. 
In the main manuscript we provide mean results over the $15$ corruptions for CIFAR-10/100C (Level 5). 
For reference, in Table~\ref{tab:supp-cifar-10-100C-results-ViT} we provide results for each individual corruption for the highest severity in the CIFAR-10/100C. 
Additionally, we also provide ActMAD results for the highest severity on the ImageNet-C benchmark.

\subsection{Continuous Adaptation to Perturbations}
In the main manuscript (Table 4) we compare with CoTTA~\cite{wang2022cotta} on their continuous adaptation benchmark and provide the mean error over the $15$ corruptions (highest severity) in the CIFAR-C benchmark.
In Table~\ref{tab:supp-randomizing-corruptions} we provide detailed results, averaged over $10$ random runs, for each individual corruption for documentation purposes.

\section{Additional insights}
\label{sec:additional-insights}
Here, we provide comparisons with EATA~\cite{eata} (for image classification) and MemCLR~\cite{memclr} (for object detection), and test ActMAD in more realistic scenarios. 
\subsection{Comparison with additional baselines}
\paragraph{Image Classification:} In Table~\ref{tab:actmad-entropy} we compare our ActMAD with two other entropy based methods, EATA~\cite{eata} and TENT~\cite{wang2020tent}.
For these results, an ImageNet pre-trained ResNet-18 is adapted to ImageNet-C.
We see that the recent state-of-the-art method for the entropy based adaptation EATA, outperforms ActMAD. 
However, we find that the logit-level supervision provided by EATA and TENT is complementary to the test-time supervision provided by ActMAD. 
Thus, combining the entropy based objectives with activation distribution matching objective from ActMAD, results in substantial performance gains. 
It is also worth highlighting, that EATA (like TENT) is also reliant on minimizing the entropy of the output distribution for test-time adaptation, rendering it unsuitable for application to regression tasks~(\eg Object Detection). 
Whereas, ActMAD is free from such constraints and is task-agnostic in nature. 

\paragraph{Object Detection:} In Table~\ref{tab:actmad-memclr} we compare our ActMAD with MemClr~\cite{memclr}. 
Our ActMAD outperforms MemCLR comfortably in their evaluation setup.
To be comparable with the results reported in their paper, we re-train a Faster-RCNN~\cite{ren2015faster} implemented in the Detectron-2 framework~\cite{wu2019detectron2} (using their default settings) on the train split of the CityScapes dataset~\cite{Cordts2016Cityscapes}. 
For TTT, we adapt this pre-trained model to the FoggyCityScapes~\cite{foggycityscapes} test split by using a batchsize of 4 and a learning rate of $5\mathrm{e}{-4}$.

\subsection{ActMAD in stationary scenarios}
In practice such scenarios can be encountered where a vehicle is stationary for a certain period of time. 
For example, when a vehicle is parked, it can happen that the sensors (\eg,~camera), mounted on the vehicle is recording a similar type of data for the stationary period.  
Ideally, the test-time training algorithm should not diverge catastrophically while adapting on this similar type of data being recorded.
To test ActMAD in such a scenario,
we choose random stationary intervals while adapting a YOLOv3 object detector from KITTI-Clear $\to$ KITTI-Rain. 
To simulate the stationary intervals, we adapt on the same batch of data for this randomly chosen interval. 
These online adaptation results are plotted in Figure~\ref{fig:parking-mode}. 
We see that during the stationary intervals, the performance of ActMAD suffers a minor degradation, however, when the diverse batches of data are encountered, ActMAD recovers quickly. 
In practice, to counter the minor degradation in performance during the stationary intervals, adaptation can be limited to batches with sufficient diversity. 
One way to achieve this can be to threshold the difference between the statistics of the intermediate activation responses from the batches received in an online manner at test-time. 
\subsection{Runtime comparison with baselines}
Contrary to test-time training approaches~(\eg~\cite{wang2020tent, kojima2022cfa}) which only adapt the affine parameters at test-time,
ActMAD proposes to optimize the entire parameter vector for test-time training. 
In order to compare the runtime for adaptation at test-time, we adapt ActMAD and other baselines to a single distribution shift in CIFAR-10C and provide the results in Table~\ref{tab:run-time}.
The runtimes are on-par for ActMAD and all other approaches, except for DUA (which only adjusts the running mean and variance estimates, without backpropagation). 

{\small
\bibliographystyle{ieee_fullname}
\bibliography{main}
}
\end{document}